\documentclass{article}

\usepackage{bm}
\usepackage{caption}

\usepackage{tikz}

\usepackage{hyperref}
\usepackage{url}

\usepackage[utf8]{inputenc} 
\usepackage[T1]{fontenc}    
\usepackage{hyperref}       
\usepackage{url}            
\usepackage{booktabs}       
\usepackage{amsfonts}       
\usepackage{nicefrac}       
\usepackage{microtype}      
\usepackage{xcolor}         
\usepackage{graphicx} 
\usepackage{subcaption}
\usepackage{enumitem}

\usepackage{amsmath}
\newcommand\scalemath[2]{\scalebox{#1}{\mbox{\ensuremath{\displaystyle #2}}}}

\usepackage{microtype}
\usepackage{graphicx}
\usepackage{booktabs} 

\usepackage{hyperref}

\usepackage[accepted]{icml2025}

\usepackage{amsmath}
\usepackage{amssymb}
\usepackage{mathtools}
\usepackage{amsthm}

\theoremstyle{plain}
\newtheorem{theorem}{Theorem}[section]

\newtheorem{lemma}[theorem]{Lemma}
\newtheorem{corollary}[theorem]{Corollary}
\theoremstyle{definition}
\newtheorem{definition}[theorem]{Definition}
\newtheorem{assumption}[theorem]{Assumption}
\theoremstyle{remark}
\newtheorem{remark}[theorem]{Remark}
\newtheorem{example}[theorem]{Example}
\usepackage[textsize=tiny]{todonotes}

\icmltitlerunning{Mirror, Mirror of the Flow: How Does Regularization Shape Implicit Bias?}

\begin{document}

\twocolumn[

\icmltitle{Mirror, Mirror of the Flow: How Does Regularization Shape Implicit Bias?}



\icmlsetsymbol{equal}{*}

\begin{icmlauthorlist}
\icmlauthor{Tom Jacobs}{yyy}
\icmlauthor{Chao Zhou}{yyy}
\icmlauthor{Rebekka Burkholz}{yyy}
\end{icmlauthorlist}

\icmlaffiliation{yyy}{CISPA Helmholtz Center for Information Security, Saarbrücken, Germany}

\icmlcorrespondingauthor{Tom Jacobs}{tom.jacobs@cispa.de}

\icmlkeywords{Implicit bias, explicit regularization, weight decay, matrix sensing, LoRA, attention, mirror flow, time-dependent Legendre function}

\vskip 0.3in
]


\printAffiliationsAndNotice{}  

\begin{abstract}

Implicit bias plays an important role in explaining how overparameterized models generalize well.
Explicit regularization like weight decay is often employed in addition to prevent overfitting.
While both concepts have been studied separately, in practice, they often act in tandem. 
Understanding their interplay is key to controlling the shape and strength of implicit bias, as it can be modified by explicit regularization.
To this end, we incorporate explicit regularization into the mirror flow framework and analyze its lasting effects on the geometry of the training dynamics, covering three distinct effects: positional bias, type of bias, and range shrinking. 
Our analytical approach 
encompasses a broad class of problems, including sparse coding, matrix sensing, single-layer attention, and LoRA, for which we demonstrate the utility of our insights.
To exploit the lasting effect of regularization and highlight the potential benefit of dynamic weight decay schedules, we propose to switch off weight decay during training, which can improve generalization, as we demonstrate in experiments.

\end{abstract}

\section{Introduction}
Regularization is a fundamental technique in machine learning that helps control model complexity, prevent overfitting and improve generalization~\citep{kukačka2017regularizationdeeplearningtaxonomy}. 
We focus on the interplay between two major categories of regularization: explicit regularization and implicit bias. We introduce both concepts within a general minimization problem. 
Consider the objective function $f$: $\mathbb{R}^n \rightarrow \mathbb{R}$ to be minimized with respect to $x$:
\begin{equation}\label{intro : standard opt}
    \min_{x\in \mathbb{R}^n} f(x).
\end{equation}
In the context of explicit regularization, a penalty term $h(x)$ is incorporated into the objective function, directly preventing the learning algorithm from overfitting~\citep{goodfellow2016deep}, as follows:
\begin{equation}\label{intro : explicit reg}
    \min_{x\in \mathbb{R}^n} f(x)+\alpha h(x),
\end{equation}
where $\alpha$ controls the trade-off between the objective and the penalty. 
This approach regulates the model capacity \citep{10.5555/3540261.3542320} and encourages simpler solutions that are more likely to generalize well to unseen data~\citep{tian2022comprehensive}. 
Common explicit regularization methods include $L_1$ (LASSO) and $L_2$ (weight decay)~\citep{bishop2006pattern}. The effectiveness of explicit regularization techniques has been demonstrated across various machine learning paradigms~\citep{arpit2016regularized}, including supervised learning, unsupervised learning and reinforcement learning.

Implicit bias~\citep{gunasekar2017implicit, woodworth2020kernel, Li2022ImplicitBO, sheen2024implicit, vasudeva2024implicit, tarzanagh2024transformerssupportvectormachines,jacobs2024maskmirrorimplicitsparsification} can be considered as an inherent aspect of the model design and optimizer that does not require explicit modifications of the objective function. 
The goal of characterizing the implicit bias is to understand how overparameterization impacts the training dynamics and, consequently, model selection.
For example, in the presence of many global minima, optimization algorithms like gradient descent inherently converge towards low-norm solutions~\citep{woodworth2020kernel,pesme2021implicit}, which impacts model properties such as generalization \citep{overparamNorm} and memorization \citep{memoryAuto}.

Implicit bias is often associated with a mirror flow~\citep{karimi2024sinkhorn, Li2022ImplicitBO}, which results from a reparameterization of $f$ by setting ${x} = g({w})$, where ${w}\in M$ and $M$ is a smooth manifold. 
It is important to highlight a fundamental distinction between the explicit regularization in the original space and the mirror flow with the objective function, formulated as follows:
\begin{equation}\label{intro : reg opt}
    \min_{{w} \in M} f(g({w})) + \alpha h({w}).
\end{equation}
The explicit regularizer $h$ now acts on the parameters ${w}$ instead of $x=g(w)$. 
Our main goal is to understand how the explicit regularization $h(w)$ affects the implicit bias, thereby shaping the effective regularization in the original parameter space $x$.
To achieve this, we analyze their interplay by integrating explicit regularization into the mirror flow framework. 

Typically, the nature and strength of implicit bias remain constant throughout training as they are inherently determined by the model parameterization. 
For instance, it has been shown that specific forms of overparametrization lead to low-rank or sparse solutions \citep{ arora2019implicitregularizationdeepmatrix, pesme2021implicit, sheen2024implicit, woodworth2020kernel, gunasekar2017implicit}, revealing a bias towards sparsity in particular settings. Nevertheless, factors such as small initialization, large learning rates and noise are needed to obtain this sparsity bias, without guarantees.
However, the inherent bias can degrade performance if it does not fit to the learning task.
Our key insight to overcome this issue is that implicit bias can be adapted and controlled by explicit and potentially dynamic regularization, which induces a time-dependent mirror flow.
To analyze the resulting optimization problem within the extended mirror flow framework \citep{Li2022ImplicitBO} and obtain convergence and optimality results, we provide sufficient conditions for the reparameterization $g$ and explicit regularization $h$.  
Additionally, we characterize the regularization $h$ in terms of $g$ to understand their interplay and impact on the Legendre function, which is associated with the implicit bias.

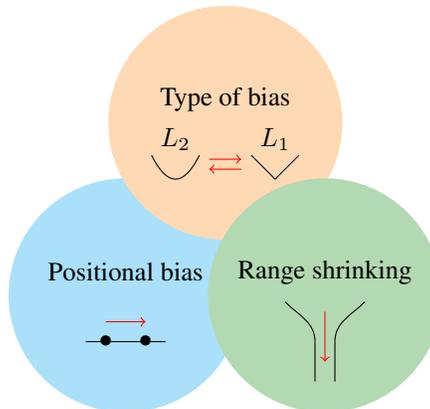
\begin{figure}
    \centering
\begin{tikzpicture}[scale = 0.53]
\begin{scope}
    \node[fill=cyan!30,circle, inner sep=1.1cm,yshift=0.75cm] at (0,0) {};
    \node at (0,2) {Positional bias};
    \node (pbA) at (-0.5,0.25) {$\bullet$};
    \node (pbB) at (0.5,0.25) {$\bullet$};
    \draw (1,0.25) -- (-1,0.25);
    \draw[->,red] (-0.5,0.75) -- (0.5,0.75); 
\end{scope}

\begin{scope}[xshift=2.5cm,yshift=4.33cm]
    \node[fill=orange!30,circle, inner sep=1.1cm,yshift=0.75cm] at (0,0) {};
    \node at (0,2) {Type of bias};
    \draw[xshift=-1.25cm,scale=0.6,domain=-1:1,smooth, variable=\x] plot ({\x}, {\x*\x});
    \draw[xshift=1.25cm,scale=0.6,domain=-1:1,smooth, variable=\x] plot ({\x}, {abs(\x)});
    \node at (-1.25,1) {$L_2$};
    \node at (1.25,1) {$L_1$};
    \draw[->,red] (0.4,0.25) -- (-0.4,0.25); 
    \draw[<-,red] (0.4,0.5) -- (-0.4,0.5); 
\end{scope}

\begin{scope}[xshift=5cm]
    \node[fill=green!50!black!30,circle, inner sep=1.1cm,yshift=0.75cm] at (0,0) {};
    \node at (0,2) {Range shrinking};
    \draw (1,1.25) to[in=90,out=180+45] (0.25,0.25);
    \draw (-1,1.25) to[in=90,out=-45] (-0.25,0.25);
    \draw (0.25,0.25) -- (0.25,-0.75);
    \draw (-0.25,0.25) -- (-0.25,-0.75);
    \draw[<-,red] (0,-0.25) -- (0,1);
\end{scope}
\end{tikzpicture}
\caption{Illustration of three established effects of explicit regularization ($\color{red}{\rightarrow}$) on implicit bias.}
    \label{fig:graphics represtation}
\end{figure}

Concretely, we identify three distinct effects: 
\begin{itemize}
    \item Type of bias: the explicit regularization changes the shape of the Legendre function and thus the implicit bias. For example, the shape changes from an $L_2$ norm to $L_1$ norm. 
    \item Positional bias:
in the standard case without explicit regularization, the global minimum of the Legendre function corresponds to the parameter initialization \citep{Li2022ImplicitBO}.
Explicit regularization shifts this global minimum, gradually moving it closer to zero during training. 
    \item Range shrinking: the explicit regularization shrinks the range of the attainable values for the Legendre function. For example,the $L_1$ norm of the parameters becomes stationary during training.
\end{itemize}
The three effects are illustrated in Figure.~\ref{fig:graphics represtation}. 
They all have a lasting impact on implicit bias, as they change the geometry of the training dynamics.

Weight decay provides an illustrative example of an explicit regularization that has a desirable impact on the training dynamics \citep{dangelo2024needweightdecaymodern}.
While \citep{10.5555/3540261.3542320} studied the effect of constant regularization on model capacity, and \citep{khodak2022initializationregularizationfactorizedneural, kobayashi2024weightdecayinduceslowrank} empirically observed that weight decay leads to sparsity bias for quadratic reparameterizations, our focus lies on understanding the effects of both constant and dynamic explicit regularization on the training dynamics and implicit bias. 
This offers a deeper theoretical insight into the interplay between implicit bias and explicit regularization.

Our theoretical framework has multiple application-relevant implications.
As examples, we explore sparse coding, matrix sensing, attention mechanisms in transformers, and Low-Rank Adaptation (LoRA) through experiments.
By switching off weight decay during training, we demonstrate the positive impact of dynamic regularization on generalization performance and analyze its effect on implicit bias.

\textbf{Contributions:}
\begin{itemize}[leftmargin=1em]
\item We establish sufficient conditions for incorporating different types of explicit regularization into the mirror flow framework and characterize their effects, focusing on three key impacts on implicit bias: positional bias shift, type of bias, and range shrinking.
\item We propose a systematic procedure for identifying appropriate regularizations and establish general convergence and optimality results. These results provide guidance on how to manage the above effects by adjusting the explicit regularization.
\item We gain the insight that explicit regularization controls the strength of implicit sparsification and has a lasting effect by changing the geometry of the training dynamics.
\item We highlight the positive impact of dynamic regularization and the resulting implicit bias through experiments such as sparse coding, matrix sensing, attention mechanism, and LoRA in fine-tuning large language models.
\end{itemize}

\section{Related Work}
\paragraph{Regularization}
There are multiple ways to regularize  in machine learning. Some of the most widely used techniques include weight decay~\citep{dangelo2024needweightdecaymodern, krogh1991simple}, data augmentation~\citep{cubuk2020randaugment, pmlr-v206-orvieto23a}, dropout~\citep{srivastava2014dropout}, and batch normalization~\citep{ioffe2015BN}. Weight decay, or $L_2$ regularization, discourages large weights to mitigate overfitting and induces a desirable change in training dynamics. This change can be effectively captured using the time-dependent mirror flow framework that we extend. 
As another example, dynamic weight decay has been proposed for ADAM to keep the gradient norms in check \citep{Xie2020OnTO}. 
In comparison, we analyze the effect of more general dynamic regularization on the implicit bias of gradient flow.

\paragraph{Implicit Bias}
The implicit bias is a well-studied phenomenon~\citep{woodworth2020kernel, gunasekar2017implicit, gunasekar2020characterizing, Li2022ImplicitBO}, which has primarily been characterized within the mirror flow framework, a well-established concept in convex optimization~\citep{Alvarez_2004, BECK2003167, Rockafellar1970ConvexA, Boyd2009ConvexO}, which we extend by introducing explicit regularization that can induce a time-dependent Legendre function.
Moreover, for convergence guarantees the time-dependent Legendre function needs to satisfy additional assumptions, i.e., it needs to be a time-dependent Bregman function.
For this class of functions we show convergence with decaying regularization.
A mirror flow can be interpreted as a gradient flow on a Riemannian manifold \citep{Li2022ImplicitBO, Alvarez_2004}, which has also been studied in stochastic gradient descent (SGD) \citep{pesme2021implicit, Even2023SGDOD, lyu2023implicit} context. 
The main observation is that large learning rates and stochastic noise from SGD have a generalization benefit by inducing sparsity, although uncontrollable.
We derive a similar but controllable benefit of explicit regularization.
Still, it is possible to also combine stochastic noise and a large learning rate with our framework.
Discrete versions of mirror flow \citep{sun2022mirror} have led to novel algorithmic designs \citep{pmlr-v130-raj21a, gonzalez2024mirrordescentalgorithmsnearly,azizan2022explicit}.
Time-dependent mirror descent, in comparison, is under-explored, except for an analysis of its intrinsic properties and an application to continuous sparsification~\citep{ radhakrishnan2021linear, jacobs2024maskmirrorimplicitsparsification}. 
Our framework covers multiple application relevant architectures and more general cases.

\paragraph{Applications of the Mirror Flow Framework}
The mirror flow framework has been applied to various architectures, including attention mechanisms in transformers \citep{vaswani2017attention, vasudeva2024implicit, sheen2024implicit, tarzanagh2024transformerssupportvectormachines, Julistiono2024OptimizingAW, pesme2024implicit}, matrix factorization \citep{li2021resolvingimplicitbiasgradient, gunasekar2017implicit, gunasekar2020characterizing} and diagonal linear networks \citep{Li2022ImplicitBO, pesme2021implicit, woodworth2020kernel}.
The implicit bias of deep matrix factorization has also been analyzed with gradient flow methods \citep{Marion2024DeepLN, arora2019implicitregularizationdeepmatrix}. Accordingly, the flow tends to be implicitly biased towards solutions with low rank or nuclear norm. We show that dynamic explicit $L_2$ regularization can further enhance this effect in the context of quadratic overparameterization.
This is illustrated through experiments on transformer networks. Moreover, we identify the inherent bias of LoRA~\citep{hu2021loralowrankadaptationlarge, wan2024efficientlargelanguagemodels} and delve into the associated challenges.
This is especially of interest, as LoRA
 has gained significant popularity in the field of large language models (LLMs), as it allows for cost-effective finetuning.
Another application is sparse coding (SC) which is similar to diagnonal linear networks with explicit regularization. This representation technique is widely employed in signal processing and pattern recognition~\citep{zhang2015survey}.
The core principle of SC is to find a sparse representation by imposing constraints, typically using the $L_0$-norm. However, this formulation leads to an NP-hard problem~\citep{Tropp04}. An alternative strategy relaxes the constraint, transforming the original problem into a convex, albeit non-smooth optimization task. Proximal algorithms have proven effective to solve these non-smooth problems~\citep{daubechies2004iterative}.  
Similarly, reparameterization with explicit regularization can be used to solve this.



\section{Theory: Integrating Explicit Regularization in the Extended Mirror Flow Framework}\label{section : theory}
We begin by reviewing the theoretical background on reparameterizations and when they induce mirror flows. Our main result, Theorem \ref{Time dep mirror : main theorem}, integrates explicit regularization into the mirror flow framework. Building on this, we explore key implications, including a geometric interpretation of the interaction between implicit bias and explicit regularization.
Using this interpretation and assuming the Legendre function $R$ is a Bregman function, we extend convergence results to the time-dependent setting by introducing the contracting property (Definition \ref{def : Bregman}, Theorem \ref{thm : convergence Bregman}).
We also prove optimality in underdetermined linear regression (Theorem \ref{main text : opt thm}).
To apply our theory in practice, we show how to choose an explicit regularizer $h$ for a given reparameterization $g$, often determined by neural network design. We characterize $h$ for known reparameterizations \cite{woodworth2020kernel, pesme2021implicit, gunasekar2017implicit} and examine their practical effects—type of bias, positional bias, and range shrinking—in Section \ref{section : effect regularization}. 

\subsection{Preliminaries}
To analyze the impact of regularization on the training dynamics of deep neural networks, we start from the gradient flow for our general optimization problem in Eq.~(\ref{intro : standard opt}).
We assume $f \in C^1(\mathbb{R}^n , \mathbb{R})$ to be a continuously differentiable objective function.
The corresponding gradient flow is:
\begin{equation*}
    dw_t = -\nabla_w f(g(w_t)) dt, \qquad w_0 = w_{init},
\end{equation*}
where $\nabla_w$ is the gradient with respect to $w$ and $g \in C^1(M, \mathbb{R}^n)$.
For a specific choice of $g$, reparameterizing the loss function $f$ leads to a mirror flow with a related implicit bias. 
For this we recall two definitions that characterize the parameterization. Moreover, we define the Legendre function needed  to define the mirror flow.

\begin{definition}\label{Definition : Regular}(Regular Parameterization, Definition 3.4 \citep{Li2022ImplicitBO})\label{def : regular}
    Let $M$ be a smooth submanifold of $\mathbb{R}^D$. A regular
parameterization $g : M \rightarrow \mathbb{R}^n$
is a $C^1$ parameterization such that the Jacobian $\partial G(w) $ is of rank $n$ for all $w \in M$.
\end{definition}
This ensures that the gradient flow for $x_t = g(w_t)$ does not have an additional null space i.e. the gradient flow can not get stuck due to reparameterization. 
For the second definition, we first need to define the Lie bracket operator where $\partial$ is the Jacobian operator.
\begin{definition}(Lie Bracket, Definition 3.4 \citep{Li2022ImplicitBO})
    Let $M$ be a smooth submanifold of $\mathbb{R}^D$. Given two $C^1$ vector fields
$X, Y$ on $M$, we define the Lie Bracket of $X$ and $Y$ as $[X, Y ](w) := \partial Y (w)X(w) - \partial X(w)Y (w)$.
\end{definition}

\begin{definition}(Commuting Parameterization, Definition 4.1 \citep{Li2022ImplicitBO})\label{def : commuting}
Let $M$ be a smooth submanifold of $\mathbb{R}^D$. A $C^2$ parameterization $g : M \rightarrow \mathbb{R}^d$ is commuting in a subset $S \subset M$ iff for any $i,j \in [n]$, the Lie bracket $\big[ \nabla g_i, \nabla g_j\big] (w) = 0$ for all $w \in S$. Moreover, we call $g$ a commuting parameterization if it is commuting in $M$. 
\end{definition}

Definition \ref{def : commuting} ensures appropriate eigen basis alignment. We now introduce the Legendre function which governs the mirror flow dynamics.

\begin{definition}(Legendre Function, Definition 3.8 (\citep{Li2022ImplicitBO})) \label{definition : Legendre function}
Let $R : \mathbb{R}^d \rightarrow \mathbb{R} \cup \{\infty\}$ be a differentiable
 convex function. We say $R$ is a Legendre function when the following holds:
 \begin{itemize}
     \item $R$ is strictly convex on $\text{int} (\text{dom} R)$.
     \item For any sequence $\{x_i\}^{\infty}_{i = 1}$ going to the boundary of $\text{dom} R$, $\lim_{i\rightarrow \infty}||\nabla R(x_i)||_{L_2}^2 = \infty$.
 \end{itemize}
\end{definition}

Appendix \ref{appendix implicit} summarizes the main aspects of the mirror flow framework \citep{Li2022ImplicitBO}, which explains their relationship.
Formally, let the reparameterization $g$ be regular (Definition \ref{def : regular}), commuting (Definition \ref{def : commuting}) and satisfy Assumption \ref{Regularization Parameterization : Ass 3.5}. 
Then, by Theorem \ref{Regularization Parameterization : Thm 4.9}, there is an Legendre function $R:$ $\mathbb{R}^n \rightarrow \mathbb{R}$ (Definition \ref{definition : Legendre function}) that follows the dynamics: 
\begin{equation}\label{intro : mirrorflow s}
    d\nabla_x R(x_t) = - \nabla_x f(x_t) dt, \qquad x_{0} = g(w_{init}). 
\end{equation} 
The Legendre function is associated with the implicit bias of the optimization. 
For example, $R$ can be the hyperbolic entropy studied in \citep{pesme2021implicit, woodworth2020kernel, Wu2021ImplicitRI}.
Depending on the initialization of the reparameterization, the hyperbolic entropy is equivalent to either $L_2$ or $L_1$ implicit regularization. 
A Legendre function $R$ that resembles an $L_1$ regularization is associated with the so-called feature learning regime, which has been argued to improve generalization performance. 
Accordingly, it presents a positive impact of overparameterization on deep learning.

Notably, in the presence of explicit regularization, the Legendre function $R$ can change over time, which has been recognized by \citet{jacobs2024maskmirrorimplicitsparsification} with the specific goal to exploit the implicit bias for gradual sparsification. 
\citep{lyu2024dichotomyearlylatephase} has analyzed how small constant weight decay impacts implicit bias to study its effect on Grokking, but has done so outside of the mirror flow framework.
We allow for dynamic and possibly large regularization of relatively general form.
While the implicit bias can change dynamically also for different reasons like large learning rate and stochastic noise as in  \citep{pesme2021implicit, lyu2023implicit}, we focus on dynamic explicit regularization to control this change. 

\subsection{Main Result}
We characterize the interplay between explicit regularization and implicit bias by a time-dependent Legendre function.
In the setting of Eq.~(\ref{intro : reg opt}) with reparameterization $g \in C^1(M, \mathbb{R}^n)$ and explicit regularization $h \in C^1(M, \mathbb{R})$, we allow the regularization strength $\alpha$ to vary over time during the gradient flow, as indicated by an index $\alpha_t$. This induces the following gradient flow:
\begin{equation*}
    dw_t = -\left(\nabla_w f(g(w_t)) + \alpha_t \nabla_w h(w_t)\right)dt, \quad w_0 = w_{init}.
\end{equation*}
To rigorously define the corresponding time-dependent mirror flow, we  define a parameterized Legendre function based on Definition \ref{definition : Legendre function}.

\begin{definition}\label{exp reg : Legendre def}
    Let $A$ be a subset of $\mathbb{R}$.
    A parameterized Legendre function is $R_{a}: \mathbb{R}^n \rightarrow \mathbb{R}^n$ such that for all $a\in A$, $R_{a}$ is a Legendre function (Definition \ref{definition : Legendre function}).
\end{definition}

The next theorem is our main result and builds on Definition~\ref{exp reg : Legendre def} and Theorem \ref{Regularization Parameterization : Thm 4.9} in the appendix. 
\begin{theorem}\label{Time dep mirror : main theorem}
    Let $(g,h)$: $M \rightarrow \mathbb{R}^{n+1}$ be regular and commuting reparameterization satisfying Assumption \ref{Regularization Parameterization : Ass 3.5}. Then there exists a time-dependent Legendre function $R_{a}$ such that 
    \begin{equation}\label{timedep implicit bias mirror flow}
        d \nabla_x R_{a_t}(x_t) = -\nabla_x f(x_t) dt, \qquad x_0 = g(w_{init}),
    \end{equation}
    where $a_t = -\int_0^t \alpha_s ds$.
    Moreover, $R_{a_t}$ only depends on the initialization $w_{\text{init}}$ and the reparameterization $g$ and regularization $h$, and is independent of the loss function $f$.
\end{theorem}
Proof. See Theorem~\ref{Time dep mirror : main theorem appendix} in the appendix.
The main steps of the proof are: 1) We apply Theorem 4.9 \citep{Li2022ImplicitBO} to the time-dependent loss function $L_t\left(x,y\right) = f\left(x\right) + \alpha_t y$ with the reparameterization $x = g(w)$ and explicit regularization $y = h(w)$ to get the resulting mirror flow with Legendre function $R(x,y)$.
2) $R$ is strictly convex. We utilize this to show that $y \rightarrow \partial_y R(x,y)$ is invertible.
3) We use the fact that the mirror flow for $y_t$ is defined by $\partial_y R(x_t,y_t) = a_t$, where $a_t = - \int_0^t \alpha_s ds$.  Next, we plug in the inverse $y_t = Q(x_t, a_t)$ into $\nabla_x R(x_t,y_t)$ to get an expression for the gradient of the time-dependent Legendre function $R$. This leads to an equation for the time-dependent mirror flow $\nabla_x R(x_t,Q(x_t, a_t)) = \mu_t$, where $\mu_t = - \int_0^t \nabla_x f(x_s) ds$.
4) In the final step, we show that $\nabla_x R\left(x, Q(x,a)\right)$, where $\nabla_x$ is the derivative with respect to the first entry, is the gradient of a Legendre function for $a$ fixed.

Theorem \ref{Time dep mirror : main theorem} characterizes the training dynamics of reparameterization with regularization. This leads to an additional geometric interpretation, which we use next.

\subsection{Geometric Interpretation}
A mirror flow can be interpreted as a gradient flow on a Riemannian manifold \citep{Li2022ImplicitBO, Alvarez_2004}.
If a Legendre function $R$ induces a mirror flow, the iterates $x_t$ follow the dynamics:
\begin{equation}\label{mirror riemann}
    d x_t = - \left(\nabla^2_x R(x_t) \right)^{-1} \nabla_x f(x_t) dt \qquad x_0 = g(w_{init}),
\end{equation}
where the manifold metric is given by $\left(\nabla^2_x R\right)^{-1}$.
Accordingly, Theorem \ref{Time dep mirror : main theorem} leads to a new geometric interpretation of a regularization. 
Specifically, $x_t$ evolves as follows:
\begin{equation}\label{exp reg : geom interpretation}
dx_t = -\left(\nabla^2_x R_{a_t}(x_t)\right)^{-1} \left(\nabla_x f(x_t) + \alpha_t \nabla_x y_t\right) dt,
\end{equation}
with initialization $x_0 = g(w_{init}) \text{ and } y_0 = h(w_{init})$, where $y_t$ is defined as in Theorem \ref{Time dep mirror : main theorem}.
Thus, the effect of regularization on the training dynamics is described by a changing Riemannian metric, where 
the metric evolves 
according to the time-dependent Legendre function $R_{a_t}$.

In practice, we can steer $a_t$ and thus influence the manifold.
Another perspective on this is that the effect of explicit regularization is stored in the time-dependent Legendre function $R_{a_t}$. 
Therefore, explicit regularization has a lasting effect on the training dynamics, even after it has been turned off, for instance.
This creates a novel connection between explicit regularization and implicit bias. 
Also past regularization influences future implicit bias by shaping the geometry.

\paragraph{Convergence}
The geometric interpretation not only provides valuable intuition but also helps us to show convergence of the mirror flow for time-dependent Bregman functions $R_{a_t}$. A Bregman function is defined as follows:

\begin{definition}(Bregman function, Definition 4.1 \citep{Alvarez_2004})\label{definition : Bregman function}
 A function $R$ is called a
 Bregman function if it satisfies the following properties:
 \begin{itemize}
     \item $\text{dom} R$ is closed. $R$ is strictly convex and continuous on $\text{dom} R$. $R$ is $C^1$ on $\text{int} (\text{dom}R ))$.
     \item For any $x \in \text{dom} R$ and $\gamma \in \mathbb{R}$,\\ $\{y \in \text{dom} R | D_R(x,y) \leq \gamma\}$ is bounded.
     \item For any $x \in \text{dom} R$ and sequence $\{x_i\}^{\infty}_{i=1} \subset \text{int}(\text{dom} R)$ such that $\lim_{i \rightarrow \infty} x_i = x$, it holds that $\lim_{i\rightarrow \infty} D_R(x,x_i) \rightarrow 0$.
     
 \end{itemize}
\end{definition}

Using Definition \ref{definition : Bregman function}, we can define the parameterized Bregman function next: 
\begin{definition}\label{def : Bregman}
    Let $A$ be a subset of $\mathbb{R}$.
    A parameterized Bregman function is $R_{a}: \mathbb{R}^n \rightarrow \mathbb{R}^n$ such that for all $a\in A$, $R_{a}$ is a Bregman function (Definition \ref{definition : Bregman function}). Furthermore, $R_a$ is called contracting if ${dR_a}/{d a} \leq 0$ for $a \in A$.
\end{definition}
\begin{remark}\label{conv remark}
 If $\alpha_t =0$ for $t \geq T$ (for a $T > 0$), we recover a gradient flow with Riemannian metric $\left(\nabla^2_x R_{a_T}\right)^{-1}$.
\end{remark}
This implication is useful for proving our next result, which highlights under which conditions we can obtain convergence if we switch off regularization at some point during training. 
We will also verify later in our experiments that this is a promising dynamic regularization strategy. 
Our next Theorem \ref{thm : convergence Bregman} gives us the convergence we are looking for by using the newly defined contracting property above.
\begin{theorem}\label{thm : convergence Bregman}
    Consider the same setting as in Theorem \ref{Time dep mirror : main theorem}. 
    Furthermore, assume that $\alpha_t \geq 0$ and $\alpha_t =0$ for all $t \geq T$, where $T > 0$. Moreover, for $a \in [b, 0]$, $R_{a}$ is a contracting Bregman function for some $b < 0$. Assume that for all $t \geq 0$ the integral $a_t : = -\int_0^t \alpha_s ds \geq b$. 
    For the loss function assume that $\nabla_x f$ is locally Lipschitz and $\text{argmin} \{ f(x) : x \in \text{dom} R_{a_{\infty}} \} $ is non-empty. Then the following holds: If $f$ is quasi-convex, $x_t$ converges to a point $x_*$ which satisfies $\nabla_x f(x_*)^T \left(x - x_*\right) \geq 0$ for $x \in dom R_{a_{\infty}}$. 
    Furthermore, if $f$ is convex, $x_*$ converges to a minimizer $f$ in the closed domain $\overline{\text{dom} R_{a_{\infty}}}$. 
\end{theorem}
Proof, see Theorem \ref{thm : convergence Bregman appendix} in the appendix.
The proof consists of two parts: a)
We show that the iterates are bounded up to time $T$ using the contracting property and quasi-convexity. b) We establish convergence after time $T$ using the geometric interpretation of the evolution of $x_t$.

\paragraph{Optimality}
To show optimality, we need more assumptions on the problem.
As common in the context of mirror flows \citep{Li2022ImplicitBO,jacobs2024maskmirrorimplicitsparsification}, we recover under-determined linear regression, as follows. 
Let $\{(z_i,y_i)\}_{i=1}^n \subset \mathbb{R}^d \times \mathbb{R}$ be a dataset of size $n$.
 Given a reparameterization $g$ with regularization $h$, the output of the linear model on the $i$-th data is $z_i^T g(w)$. The goal is
 to solve the regression for the target vector $Y = (y_1,y_2,\hdots,y_n)^T$ and input vector
 $Z = (z_1,z_2,\hdots,z_n)$.

\begin{theorem}\label{main text : opt thm}
    Assume the same setting as Theorem \ref{Time dep mirror : main theorem}. 
    Furthermore, assume that $\alpha_t \geq 0$ and $\alpha_t =0$ for all $t \geq T$, where $T > 0$.
    If $x_t$ converges when $t \rightarrow \infty $ and the limit $x_{\infty} = \lim_{t \rightarrow \infty} x_t$ satisfies $Z x_{\infty} = Y$, then the gradient flow minimizes the changed regularizer $R_{a_T}$:
    \begin{equation}\label{equation : opt}
        x_{\infty} = \text{argmin}_{x: Z x = Y} R_{a_T}(x).
    \end{equation} 
\end{theorem}
Proof. See Theorem \ref{appendix : opt thm}.

This theorem extends known optimality results on matrix sensing \citep{gunasekar2017implicit, Wu2021ImplicitRI} and diagonal linear networks (sparse coding) \citep{pesme2021implicit, woodworth2020kernel, jacobs2024maskmirrorimplicitsparsification}. 


\subsection{Characterization of the Explicit Regularization}
To make use of the established theoretical results in practice and develop promising regularization strategies, we characterize the explicit regularization $h$ for two important classes of reparameterizations: separable and quadratic reparameterizations.

\paragraph{Separable Reparameterizations}
Our next result encompasses most previously studied reparameterizations within mirror flow framework\cite{woodworth2020kernel, pesme2021implicit, gunasekar2017implicit}.

\begin{corollary}\label{corollary : seperable}
    Let $g$ be a separable reparameterization such that $g_i(w_i)= \sum_{j = 1}^{m_i} g_{i,j}(w_{i,j})$ and $h(w) = \sum_{i =1}^n\sum_{j =1}^{m_i}  h_{i,j}(w_{i,j})$, where $g_{i,j} : \mathbb{R} \rightarrow \mathbb{R}$ and $h_{i,j} : \mathbb{R} \rightarrow \mathbb{R}$.
    Furthermore, assume that $g$ and $h$ are analytical functions. Then, if and only if $h$ and $g$ satisfy
    \begin{equation*}
        h_{i,j} = c_{i,j} g_{i,j} \qquad \forall i \in [n],j \in [m_i],
    \end{equation*}
    where $c_{i,j} \in \mathbb{R}$ is a constant, Theorem \ref{Time dep mirror : main theorem} applies.
\end{corollary}
Proof. The result follows from the commuting relationship between $g$ and $h$. We use that the Wronskian between two analytical functions is zero if and only if they are linearly dependent \citep{Bcher1901CertainCI}.

The next two examples highlight the utility of Corollary \ref{corollary : seperable}. 
\begin{example}\label{ex1 pesme}
    The reparameterization $g: \mathbb{R}^n \times \mathbb{R}^n \rightarrow \mathbb{R}^n$ such that $g(u,v) = u^2 -v^2$ with regularization of the form $h(u,v) = \sum_{i=1}^n c_u u^2_i - c_v v^2_i$. Setting $c_u = 1$ and $c_v = -1$ leads to weight decay regularization on the reparameterization. 
\end{example}

Example \ref{ex1 pesme} has also been used to study the effect of stochasticity on overparameterized networks \cite{pesme2021implicit}.
We present a more general class of examples that always results in a well-posed optimization problem, i.e., $h$ is positive.

\begin{example}
    Consider the reparameterization $g: \mathbb{R}^n \times \mathbb{R}^n \rightarrow \mathbb{R}^n$ such that  $g(u,v) = a(u) - b(v)$, where $a$ and $b$ are positive, analytical, increasing functions. Then, the regularization $ \sum_{i =1}^n c_u a_i(u) - c_v b_i(v)$ can always be employed.
    By selecting $c_u \geq 0$ and $c_v \leq 0$, the optimization problem remains well-posed.
\end{example}
To give concrete examples, this approach encompasses $u^{2k} - v^{2k}$ \citep{woodworth2020kernel} and $\log  u - \log  v$. 

\paragraph{Quadratic Reparameterizations}
Next, we will discuss the class of quadratic reparameterizations, as described in Theorem 4.16 in \citep{Li2022ImplicitBO}.

\begin{theorem}\label{thm : quadratic param}
    In the setting of Theorem \ref{thm : convergence Bregman}, consider the commuting quadratic parametrization $G$: $\mathbb{R}^D \rightarrow \mathbb{R}^d$ and $H$: $\mathbb{R}^D \rightarrow \mathbb{R}$, where each $G_i(w) = \frac{1}{2} w^T A_i w$ and $H(w) = \frac{1}{2} w^T B w$ with symmetric matrices $A_1, A_2, \hdots, A_d \in \mathbb{R}^{D\times D}$ and symmetric matrix $B\in \mathbb{R}^{D\times D}$ that commute with each other, i.e., $A_i A_j - A_j A_i = 0$ for all $i, j \in [d]$ and $B A_j - A_j B = 0$ for all $ j \in [d]$. For any $w_{init} \in \mathbb{R}^D$, if
${\nabla_w G_i(w_{init})}^d_{i=1} = {A_i w_{init}}_{i = 1}^d$ and ${\nabla_w H(w_{init})} = {B w_{init}}$ are linearly independent, then the following holds:

1) $Q_a(\mu)  = \frac{1}{4}||\exp (a B + \sum_{i=1}^d \mu_iA_i)w_{init}||^2_{L_2}$ is a time-dependent Legendre function with domain $\mathbb{R}^{d}$.

2) For all $a \in \mathbb{R} $, $R_{a}$ is Bregman function with $\text{dom} R_{a} = \overline{\text{range}\nabla_x Q_a}$. Furthermore, if $B$ is positive semi-definite, then $\frac{d R_a}{da} \leq 0$, therefore Theorem \ref{thm : convergence Bregman} applies.
\end{theorem}
Proof. 
The first statement is derived by applying Theorem 4.16 from \citep{Li2022ImplicitBO}.
The second statement follows from recognizing that $\exp (a B)$ acts as a linear transformation of the initialization $w_{init}$. 
Subsequently, applying Theorem 4.16 of \citep{Li2022ImplicitBO} gives the first part of the last statement. It remains to show that $R_a$ is contracting. 
Since $B$ is positive semi-definite, it follows that $\frac{d}{da}Q_a \geq 0$. By the reverse ordering property of convex conjugation, we have that $\frac{d}{da} R_a \leq 0$. For completeness, let $h >0$, then for $a \in \mathbb{R}$, we have $Q_{a+h} \geq Q_{a}$. 
Applying the reverse ordering property implies $R_{a+h} \leq R_{a}$. Rearranging and dividing by $h$ gives $\frac{1}{h}\left( R_{a+h} - R_a\right) \leq 0$. 
Taking the limit $h \rightarrow 0$ concludes the proof.
\begin{remark}
    For the time-dependent Bregman function in Theorem \ref{thm : quadratic param} to be contracting, $B$ needs to be positive semi-definite.
\end{remark}
Theorem \ref{thm : quadratic param} encompasses recent works on the reparameterization $g(m,w) = m \odot  w$, where $\odot$ denotes pointwise multiplication (Hadamard product). 
It has been proposed to sparsify neural networks \citep{jacobs2024maskmirrorimplicitsparsification}, and extends work on matrix sensing and transformers \citep{Wu2021ImplicitRI, gunasekar2017implicit, sheen2024implicit}. 
Furthermore, $B = I$ corresponds to weight decay on the reparameterization, which is often used in practice.
Having identified classes where we can determine $h$, we next present applications to illustrate how time dependence influences the dynamics. 


\section{Analysis: Effects of Explicit Regularization}\label{section : effect regularization}
We introduce several time-dependent Legendre functions to demonstrate the wide applicability of our analysis. 
We aim to gain insights into how explicit regularization affects implicit bias during training. In particular, we focus on three distinct effects, as summarized below:
\begin{itemize}
    \item Type of bias: The shape of $R_a$
 changes with $a$.
    \item Positional bias: The global minimum of $R_a$
 changes with $a$. 
    \item Range shrinking: The range of $\nabla R_a$
 can shrink due to a specific choice of $a$.
\end{itemize}

\begin{figure}[!htb]
\centering
\begin{subfigure}[b]{0.35\textwidth}
\centering
         \includegraphics[width=0.95\textwidth]{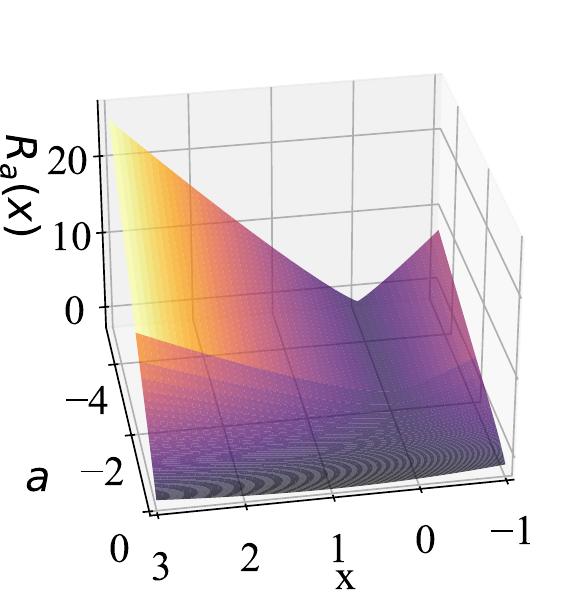}
        \caption{Time-dependent hyperbolic entropy.}
       \label{fig: Breg evol mw}
\end{subfigure}
\hfill
\begin{subfigure}[b]{0.35\textwidth}
\centering
         \includegraphics[width=0.95\textwidth]{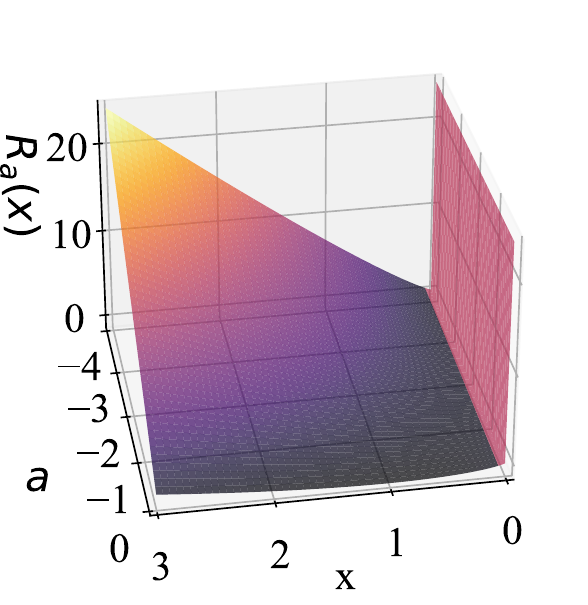}
        \caption{Time-dependent entropy.}
       \label{fig: Breg evol ent}
\end{subfigure}
\caption{Illustrations of the positional bias and type of bias effects of explicit regularization on the time-dependent Legendre function. In both figures $a = -\int_0^t \alpha_s ds$. Depending on the initialization of $m\odot w$ the time-dependent Legendre function is given by Fig~\ref{fig: Breg evol mw} or Fig~\ref{fig: Breg evol ent}. Both exhibit a type change towards $L_1$ minimization.}
\label{fig : entropies}
\end{figure}

\paragraph{The Reparameterization $m\odot w$} 
The reparameterization $g(m,w) = m \odot w$ is an exemplary quadratic reparameterization, which can also be interpreted as the spectrum of more general quadratic reparameterizations.
When the initialization satisfies $|w_{0}| < |m_{0}|$, then Theorem \ref{thm : quadratic param} holds.
We can compute the time-dependent Bregman function $R_{a}(x)$:
\begin{equation}\label{mw time}
\scalemath{0.85}{\frac{1}{4} \sum_{i=1}^d x_i \text{arcsinh}\left(\frac{x_i}{A_i(a)}\right) -\sqrt{x_i^2 + A_i(a)^2} - x_i \log \left(\frac{u_{i,0}}{v_{i,0}}\right)}
\end{equation}
where $A_i(a) : = 2 \exp(2a) u_{i,0} v_{i,0}$ and $u_0 = (m_{0} + w_{0})/\sqrt{2}$ and $v_0 =(m_{0} - w_{0})/\sqrt{2}$.
This adapts the hyperbolic entropy \citep{Li2022ImplicitBO,woodworth2020kernel, Wu2021ImplicitRI}, which now is dependent on $a$. 
Note that we used Theorem \ref{thm : quadratic param} to find $R_a$, we can invert the corresponding function $Q_a(\mu)$, where $\mu = -\int_0^t \nabla_x f(x_s) ds$.
The regularization thus affects the time-dependent Legendre function by changing $a$. This allows us to modulate between an implicit $L_2$ and $L_1$ regularization through explicit regularization \citep{jacobs2024maskmirrorimplicitsparsification}. Moreover, $a$ also controls the location of the global minimum, a smaller $a$ corresponds to moving it closer to zero. 
Therefore, we both change the type of bias and the positional bias. 
Similarly, in case $w_0 = m_0 >0$ we recover the entropy \citep{Wu2021ImplicitRI}:
\begin{equation}\label{wu : implicit bias}
    \sum_{i=1}^n \left( \text{log}\left(\frac{1}{B_i(a)}\right) - 1\right) x_i + x_i \text{log} x_i,
\end{equation}
where $B_i(a) := x_0 \exp{\left( 2a \right)} $. 
Here $a$ modulates between maximizing and minimizing the $L_1$-norm.
Note that both time-dependent Bregman functions are contracting on $a \in (-\infty, 0]$.
Moreover, Figure~\ref{fig : entropies} 
illustrates the effects of type of bias and positional bias for $m \odot w$. 

\paragraph{Quadratic Reparameterizations} 
Building on the characterization of the reparameterization $m \odot w$, we study the more general quadratic reparameterizations with weight decay ($B =I$).
\begin{table}[ht]
\caption{Quadratic parametrization.}
    \centering
    \begin{tabular}{c|c}
    Matrix sensing & $UU^T$ \\ \hline
       Attention  &  $\textit{SoftMax}( QK^T) V$ \\ \hline
       LoRA  &   $W_0 + A B$ \\
    \end{tabular}
    \label{tab:quad param}
\end{table}
This covers multiple architectures, including matrix sensing, attention and LoRA, as explained in Table \ref{tab:quad param}. For the dimensions of the parameters, see Table \ref{tab:appendix quad param} in the Appendix.
In general, however, the assumptions of Theorem \ref{thm : quadratic param} might not hold.
In case of matrix sensing, we are able to apply both Theorem \ref{thm : quadratic param} and \ref{main text : opt thm}, where the time-dependent Bregman function for the eigenvalues is given by Eq.~(\ref{wu : implicit bias}). Thus, weight decay modulates between maximizing and minimizing the nuclear norm of the matrix $X = UU^T$. 
The details are given in Appendix \ref{appendix: section opt}.
For attention, a common building block of Transformer architectures \citep{sheen2024implicit, tarzanagh2024transformerssupportvectormachines}, additional assumptions such as the alignment property are required.
It is worth noting that attention also has a value matrix $V$ and an activation function. 
Assuming $V$ is not trainable and that the function $f$ encompasses the activation function, the gradient flow dynamics of $X = QK^T$ is described by Theorem \ref{Time dep mirror : main theorem}. This characterizes the implicit bias and can be interpreted as a proxy for training full attention.
Similarly, for LoRA, a finetuning mechanism for LLMs \citep{hu2021loralowrankadaptationlarge, wan2024efficientlargelanguagemodels}, the training dynamics of $X= AB$ is described by Theorem \ref{Time dep mirror : main theorem}, assuming the alignment property in addition.

As observed in \cite{khodak2022initializationregularizationfactorizedneural} for quadratic reparameterizations, weight decay promotes small nuclear norm. 
This is not the full picture, however. 
According to our results, weight decay changes the manifold geometry according to Eq.~(\ref{exp reg : geom interpretation}), which leads to an implicit bias that modulates between the Frobenius and nuclear norm of the matrix $X$. 
The eigenvalues in Eq.~(\ref{mw time}) are therefore subject to a time-dependent Bregman function $R_{a_t}$.
This is the most accurate description of the implicit bias among the Bregman functions in Eq.~(\ref{mw time}) and Eq.~(\ref{wu : implicit bias}) that we have discussed for the parameterization $m \odot w$, taking the initializations of both attention and LoRA into account. 
In case of attention, the matrices are randomly initialized, which makes a coupling between the spectrum of $K$ and $Q$ unlikely, i.e., $m_0 = w_0$.
In case of LoRA, the initialization is $A = 0$ and $B$ is random.
Regardless of the initialization, weight decay would still regularize towards the nuclear norm, but a coupled initialization would constrain the eigenvalues to be either positive or negative.
\paragraph{The Reparameterization $u^{2k} - v^{2k}$}
The reparameterization $u^{2k} - v^{2k}$ serves as a proxy for deep neural networks \citep{woodworth2020kernel} and provides an example of range shrinking due to explicit regularization.
We consider the regularization $h(u,v) ={\sum_{i=1}^n}u_i^{2k} + v_i^{2k}$ as allowed by Corollary \ref{corollary : seperable}.
The current reparameterization also exhibits a change in the implicit bias from $L_2$ to $L_1$, shown in Theorem 3 in \citep{woodworth2020kernel}.
Unfortunately, there is no analytical formula available for the Legendre function in this case. 
Therefore we only derive the flow and its domain, which is the range of the time-dependent mirror flow.
The flow $Q_{a_t}(\mu_t)$ is given by
\begin{equation}
\scalemath{0.85}{
    d_k\left(\frac{1}{\mu_t + a_t + c_u}\right)^{\frac{2k}{2k-2}} -  d_k\left(\frac{1}{-\mu_t + a_t + c_v}\right)^{\frac{2k}{2k-2}}},
\end{equation}
where $d_k = \left(\left(2k-2\right)\left(2k\right)\right)^{\frac{2k}{2k-2}}$, $\mu_t = -\int_0^t \nabla f(x_s)ds $ and $a_t = -\int_0^t \alpha_s ds$. 
We have that $\text{dom} \nabla_x R_a = \text{int} \text{dom} Q_a $ for $a$ fixed, where int refers to the interior of the domain (see Lemma 4.8 \citep{Li2022ImplicitBO}). Furthermore, note that $\mu \in (-c_u- a, c_v + a)$.
Since $a_t$ is negative, the domain of $Q_a$ shrinks over time. 
Thus, the range of $\nabla_x R_a$ shrinks accordingly. 
This also lessens the set of acceptable solutions of the original optimization problem, which can make its solution harder.
Figure~\ref{fig: Breg evol unvn} in Appendix \ref{appendix acc} illustrates the effect of range shrinking with an approximation of the time-dependent Legendre function.

\paragraph{Other Reparameterizations}
Appendix \ref{appendix log} and \ref{appendix : other param} present more reparameterizations. 
In particular, we analyze a reparameterization that induces an $L_1$ to $L_2$ change in the type of bias (in contrast to the more prevalent flipped change from $L_2$ to $L_1$).
In addition, we highlight limitations of the framework by considering deeper reparameterizations.

\section{Experiments}\label{section : experiments}
We conduct three experiments to support our theoretical analysis. 
The first experiment on matrix sensing illustrates the positional bias and type change following Theorem \ref{main text : opt thm}. 
Accordingly, we turn off weight decay at some point during training and compare it with a linear reparameterization with $L_1$ regularization. 
The second and third experiments are finetuning a pretrained transformer network and an LLM with LoRA, respectively. 
Both exhibit a gradual change of the implicit bias from Frobenius to nuclear norm, while demonstrating a lasting effect of dynamic regularization, which leads to better generalization.
Notably, this highlights that our insights extend even to settings where our assumptions are not strictly met.
Moreover, Appendix \ref{appendix exp log} discusses the range shrinking effect on sparse coding.
Note that the change in positional bias is present in all experiments. 



\begin{figure}[!htb]
    \centering
    \includegraphics[width=0.96\linewidth]{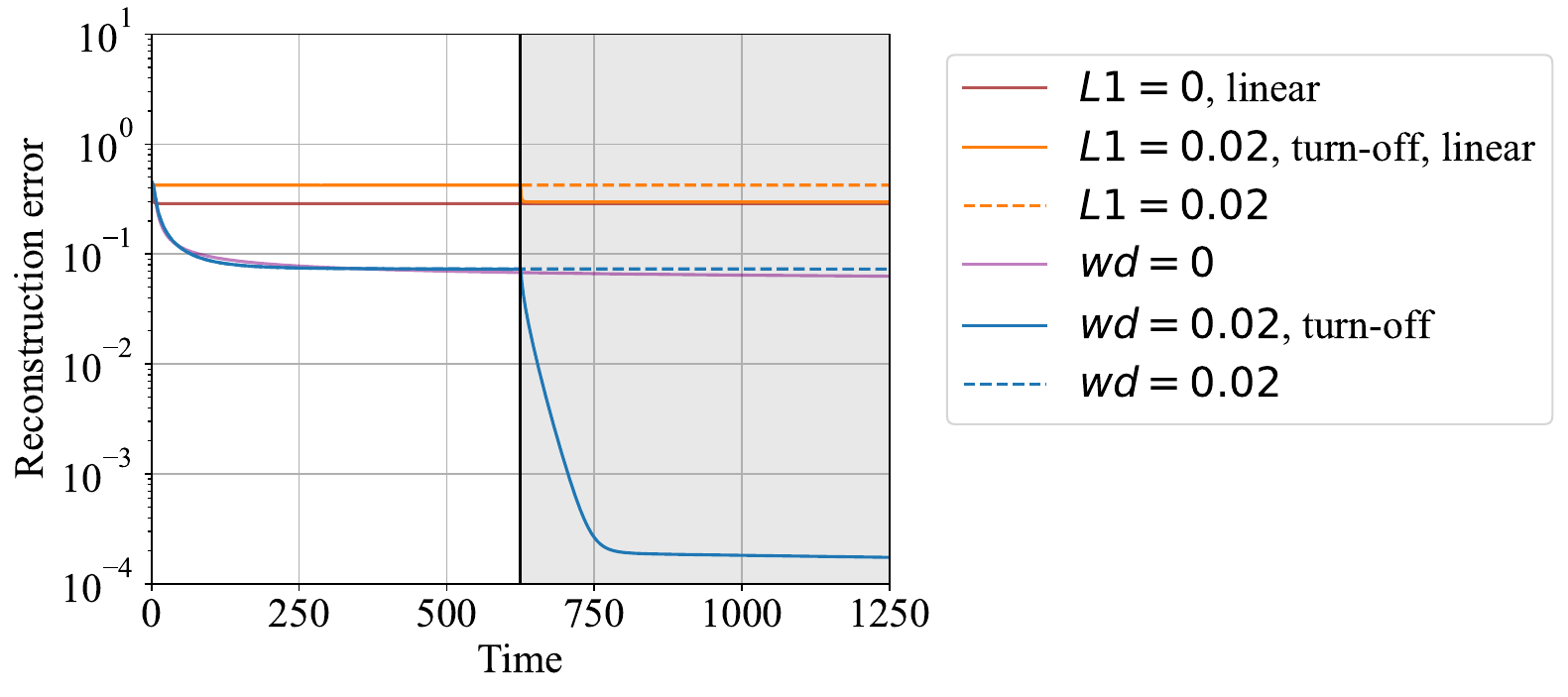}
    \caption{Recovering the sparse ground truth by turning weight decay off for matrix sensing at $T = 675$. In contrast, a linear reparameterization with $L_1$ regularization goes towards the minimal $L_2$ norm solution after a switch-off.}
    \label{fig:sensing R}
\end{figure}

\paragraph{Recovering the Sparse Ground Truth in Matrix Sensing}
We consider a matrix sensing experiment with the setup of \citep{Wu2021ImplicitRI}. 
Details can be found in Appendix \ref{appendix: section opt}. 
The ground truth is a sparse matrix $X^*$ and the reparameterization is $X = U U^T$. 
When initialized with $U_0 U^T_0 = \beta I$, the eigenvalues of $X$ satisfy Theorem \ref{main text : opt thm} with the time-dependent Legendre function in Eq.~(\ref{wu : implicit bias}). 
For experiments labeled "turn-off", we turn off the regularization at time $T = 625$.
Figure \ref{fig:sensing R} demonstrates that we recover the ground truth after turning off weight decay for the quadratic reparameterization.
This can be explained by the fact that the positional bias of the eigenvalues moves closer to zero over time, which causes the type of bias to change to the nuclear norm (see Figure \ref{fig: sensing nuc}) in the appendix.
Disabling weight decay in this context reveals the accumulated effects of regularization. 
In contrast, using constant weight decay would prevent us from exploiting this effect. 
Also note that a linear reparameterization with $L_1$ regularization could achieve high sparsity but at the expense of reconstruction accuracy, as it cannot recover the ground truth. 
The geometry of the manifold that defines the implicit bias does not allow for it.

\begin{figure}[!htb]
  \centering
    \begin{subfigure}[b]{0.45\textwidth}
    \centering
    \includegraphics[width=0.96\textwidth]{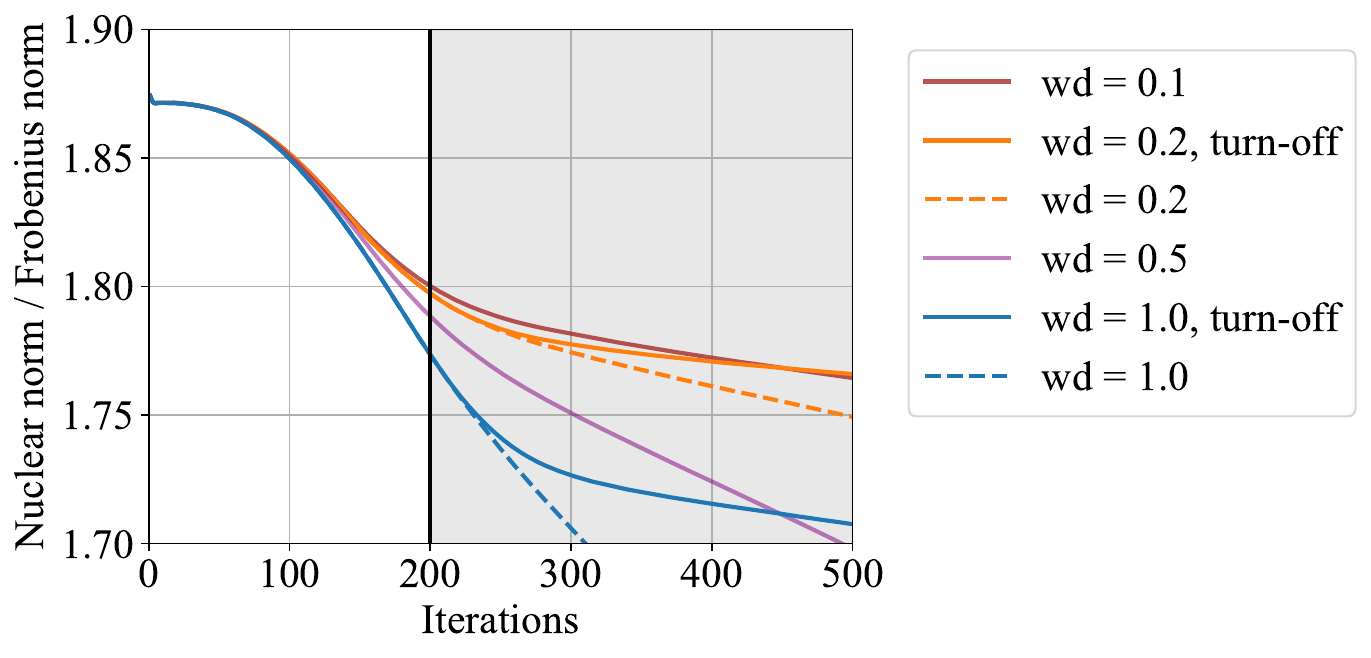}
    \caption{LoRA.}
    \label{fig:LORA_ratio}
    \end{subfigure}
    \hfill
    \begin{subfigure}[b]{0.45\textwidth}
    \centering
    \includegraphics[width=0.96\textwidth]{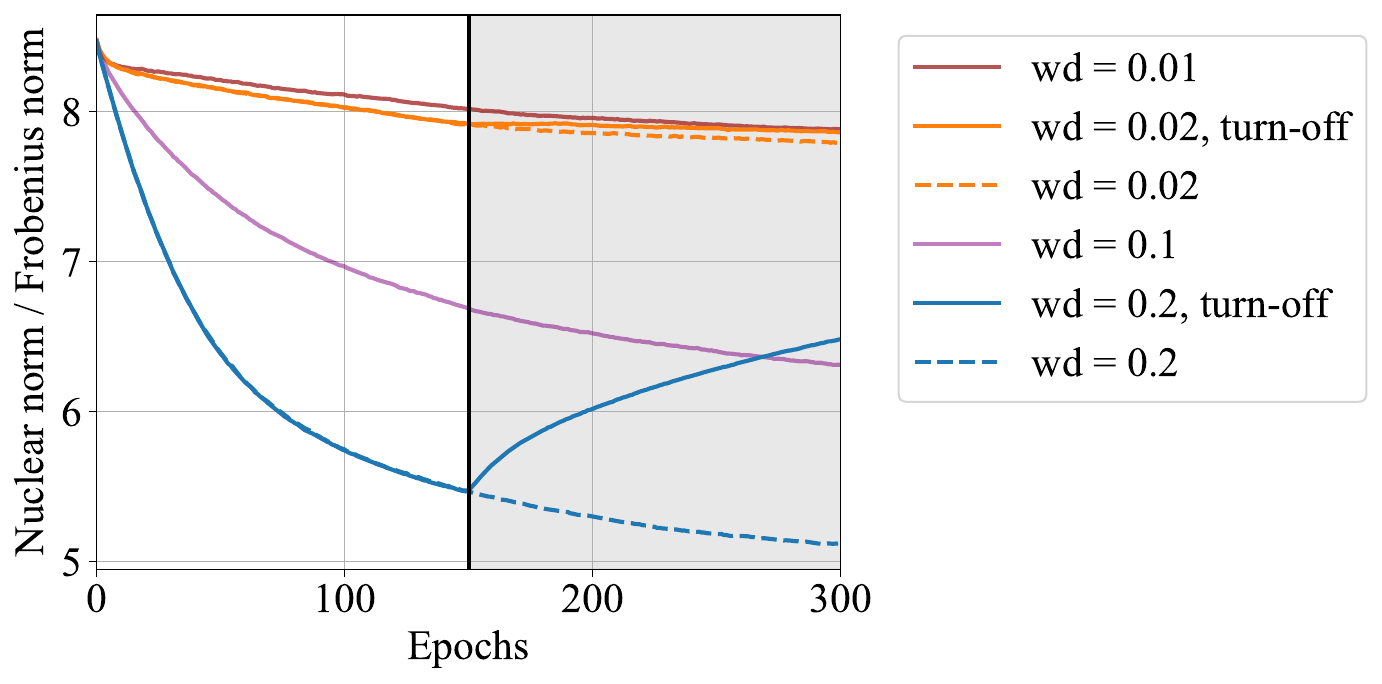}
    \caption{Attention.}
    \label{fig: ratio attention}
    \end{subfigure}
    \caption{Ratio between the nuclear norm and Frobenius norm for LoRA and attention. Training with higher weight decay and then turning it off in the shaded region allows for exploring the parameter space at higher sparsity. This creates a window opportunity to improve performance in a relatively sparse regime.}
    \label{fig: ratios}
\end{figure}

\paragraph{Turning-off Weight Decay for LoRA and Attention}
Next, we aim to track the effect of the explicit regularization on the quadratic reparameterizations in Table \ref{tab:quad param}.
For LoRA, we calculate the nuclear norm and Frobenius norm of the matrix product $X =AB$, averaging these values across all layers, and then computing their ratio. 
For each attention layer, we apply the same procedure to the product of the query and key matrices $X= QK^T$.
The ratio allows us to track the relative sparsity of the matrix.
With LoRA, we finetune GPT2 \citep{radford2019language} on the \textit{tiny\_shakespeare} \citep{karpathy} dataset, training for $500$ iterations in two different types of settings. 
In case of "turn-off", we turn the weight decay off at iteration $200$.
Furthermore, we fine-tune a pretrained ViT on ImageNet for $300$ epochs, turning the weight decay off at epoch $150$.
Figure.~\ref{fig: ratios} shows that increasing weight decay reduces the reported norm ratio, indicating a change in the type of bias from $L_2$ to $L_1$. 
Moreover, when weight decay is turned off, the ratio intersects with other ratios that are attained by constant weight decay. 
This creates a "window of opportunity" for unconstrained training with a relatively low nuclear norm, leading to improved test accuracy (see Appendix \ref{appendix acc}), in particular, in comparison with constant weight decay trajectories.
For a ViT on ImageNet this can lead to more than $1\%$ improved validation accuracy for similar relative sparsity.

\section{Discussion}
We have introduced a framework for analyzing the impact of explicit regularization on implicit bias and provided a novel geometric interpretation of their interplay. 
By extending the mirror flow framework, we have outlined a method to control dynamic implicit bias through dynamic explicit regularization. 
Our analysis has characterized their joint effects on the training dynamics, including positional bias, type of bias, and range shrinking. 
Additionally, we have established a systematic procedure for identifying suitable regularizations for given reparameterizations and established convergence and optimality within our framework.
As the implicit bias can change dynamically during training, it is associated with a time-dependent Legendre function, which might be conceptually of independent interest.
To demonstrate the utility of our theory in applications, we have presented experiments on sparse coding, matrix sensing, attention in transformers, and LoRA fine-tuning.
As we found, switching off weight decay at some point during training could improve generalization performance by exploiting the accumulated effect of past regularization. 
In future, our insights could guide the development of more effective regularization techniques that account for implicit bias, such as dynamic weight decay strategies tailored to specific model architectures. 
Moreover, our framework could be used to analyze the impact of early stopping and extended to incorporate other regularization factors like the impact of a large learning rate and stochastic noise.



\newpage

\section*{Acknowledgements}
The authors thank Celia Rubio-Madrigal for proofreading and designing Figure \ref{fig:graphics represtation}.
Moreover, the authors gratefully acknowledge the Gauss Centre for Supercomputing e.V. for funding this project by providing computing time on the GCS Supercomputer JUWELS at Jülich Supercomputing Centre (JSC). We also gratefully acknowledge funding from the European Research Council (ERC) under the Horizon Europe Framework Programme (HORIZON) for proposal number 101116395 SPARSE-ML.

\section*{Impact Statement}

This paper presents work whose goal is to advance the field of 
Machine Learning. There are many potential societal consequences 
of our work, none which we feel must be specifically highlighted here.

\bibliography{neurips}
\bibliographystyle{icml2025}

\newpage
\appendix
\onecolumn

\section{Implicit bias framework}\label{appendix implicit}
In this section, for completeness, we present the existing results for the mirror flow framework. Consider the optimization problem in Eq.~(\ref{intro : standard opt}) for a loss function $f: \mathbb{R}^n \rightarrow \mathbb{R}$
\begin{equation*}
    \min_{x \in \mathbb{R}^n} f(x).
\end{equation*}

We can use the implicit bias framework to study the effect of overparameterization.
An overparameterization can be accomplished by introducing a function $g : M \rightarrow \mathbb{R}^n$, with $M$ a smooth manifold. 
For particular $g$, the reparameterization of the loss function $f$ leads to a mirror flow.
A general framework is given in \citep{Li2022ImplicitBO} to study the implicit bias in terms of a mirror flow. 
Let $R : \mathbb{R}^n \rightarrow \mathbb{R}$ be a Legendre function (Definition \ref{definition : Legendre function}), then the mirror flow is described by 
\begin{equation}\label{intro : mirrorflow}
    d\nabla_x R(x_t) = - \nabla_x f(x_t) dt, \qquad x_{init} = g(w_{init}) 
\end{equation}
\citep{Li2022ImplicitBO} provide a sufficient condition for the reparameterization $g$ such that it induces a mirror flow Eq.~(\ref{intro : mirrorflow}). The Legendre function $R$ controls the implicit bias.

\begin{definition}(Legendre function Definition 3.8 (\citep{Li2022ImplicitBO})) \label{definition : Legendre function appendix}
Let $R : \mathbb{R}^d \rightarrow \mathbb{R} \cup \{\infty\}$ be a differentiable
 convex function. We say $R$ is a Legendre function when the following holds:
 \begin{itemize}
     \item $R$ is strictly convex on $\text{int} (\text{dom} R)$.
     \item For any sequence $\{x_i\}^{\infty}_{i = 1}$ going to the boundary of $\text{dom} R$, $\lim_{i\rightarrow \infty}||\nabla R(x_i)||_{L_2}^2 = \infty$.
 \end{itemize}
\end{definition}

In order to recover the convergence result in Theorem 4.14 in \citep{Li2022ImplicitBO} the function $R$ also needs to be a Bregman function, which we define in Definition \ref{definition : Bregman function}. 

\begin{definition}(Bregman function Definition 4.1 \citep{Alvarez_2004})\label{definition : Bregman function appendix}
 A function $R$ is called a
 Bregman function if it satisfies the following properties:
 \begin{itemize}
     \item $\text{dom}R$ is closed. $R$ is strictly convex and continuous on $\text{dom} R$. $R$ is $C^1$ on $\text{int} (\text{dom}R ))$.
     \item For any $x \in \text{dom} R$ and $\gamma \in \mathbb{R}$, $\{y \in \text{dom} R | D_R(x,y) \leq \gamma\}$ is bounded.
     \item For any $x \in \text{dom} R$ and sequence $\{x_i\}^{\infty}_{i=1} \subset \text{int}(\text{dom} R)$ such that $\lim_{i \rightarrow \infty} x_i = x$, it holds that $\lim_{i\rightarrow \infty} D_R(x,x_i) \rightarrow 0$.
     
 \end{itemize}
\end{definition}

For a reparameterization to induce a mirror flow with a corresponding Legendre function we first have to give two definitions. 
Furthermore, we define $\partial g$ as the Jacobian of the function $g$. 

\begin{definition}\label{Definition : Regular appendix}(Regular Parmeterization Definition 3.4 \citep{Li2022ImplicitBO})\label{def : regular appendix}
    Let $M$ be a smooth submanifold of $\mathbb{R}^D$. A regular
parameterization $g : M \rightarrow \mathbb{R}^n$
is a $C^1$ parameterization such that $\partial G(w) $ is of rank $n$ for all $w \in M$.
\end{definition}

For the second definition, we first need to define what a Lie bracket is.
\begin{definition}(Lie bracket Definition 3.4 \citep{Li2022ImplicitBO})
    Let $M$ be a smooth submanifold of $\mathbb{R}^D$. Given two $C^1$ vector fields
$X, Y$ on $M$, we define the Lie Bracket of $X$ and $Y$ as $[X, Y ](w) := \partial Y (w)X(w) - \partial X(w)Y (w)$.
\end{definition}

\begin{definition}(Commuting Parameterization Definition 4.1 \citep{Li2022ImplicitBO})\label{def : commuting appendix}
Let $M$ be a smooth submanifold of $\mathbb{R}^D$. A $C^2$ parameterization $g : M \rightarrow \mathbb{R}^d$ is commuting in a subset $S \subset M$ iff for any $i,j \in [n]$, the Lie bracket $\big[ \nabla g_i, \nabla g_j\big] (w) = 0$ for all $w \in S$. Moreover, we call $g$ a commuting parameterization if it is commuting in the entire $M$.
\end{definition}

Besides these two definitions, we need to make an additional assumption on the flow of the solution.
We define the solution of the gradient (descent) flow of a function $f : M \rightarrow \mathbb{R}^n$ initialized at $x \in M$
\begin{equation}\label{Regularized Parameterizations : Gradient FLow}
    d x_t = -\nabla_x f(x_t) dt \qquad x_0 = x
\end{equation}
as $x_t = \phi_x^t(x)$ which is well defined if the solution exists. Using this we can make the following assumption.

\begin{assumption}(Assumption 3.5 \citep{Li2022ImplicitBO})\label{Regularization Parameterization : Ass 3.5}
    Let $M$ be a smooth submanifold of $\mathbb{R}^D$ and $g : M \rightarrow \mathbb{R}^n$ be a reparameterization. We assume that for any $w \in M$ and $i \in [n]$, $\phi_{g_i}^t(w)$ is well-defined for $t\in (T_-, T_+)$ such that either $\lim_{t \rightarrow T_+} ||\phi_{g_i}^t(w)||_{L_2} = \infty$ or $T_+ = \infty$ and similarly for $T_-$. Also, we assume that for any $w \in M$ and $i,j \in [n]$, it holds that for $(t,s) \in \mathbb{R}^2$ that $\phi_{g_i}^s \circ \phi_{g_j}^t(w)$ is well-defined iff $\phi_{g_j}^t \circ \phi_{g_i}^s(w)$
\end{assumption}


Using these definitions we state the known result for mirror flow.

\begin{theorem}\label{Regularization Parameterization : Thm 4.9}(Theorem 4.9 \citep{Li2022ImplicitBO})
    Let $M$ be a smooth submanifold of $\mathbb{R}^D$ and $g : M \rightarrow \mathbb{R}^n$ be a commuting and regular parameterization satisfying Assumption \ref{Regularization Parameterization : Ass 3.5}. For any initialization $w_{\text{init}} \in M$, consider the gradient flow for any time-dependent loss function $L_t : \mathbb{R}^d \rightarrow \mathbb{R}$:
    \begin{equation*}
        d w_t = - \nabla_w L_t(g(w_t)) dt, \qquad w_0 = w_{\text{init}}.
    \end{equation*}
    Define $x_t = g(w_t)$ for all $t \geq 0$, then the dynamics of $x_t$ is a mirror flow with respect to the Legendre function $R$ given by Lemma 4.8 in \citep{Li2022ImplicitBO}, i.e.,
    \begin{equation*}
        d \nabla_x R(x_t) = -\nabla_x L_t(x_t) dt, \qquad x_0 = g(w_\text{init}).
    \end{equation*}
    Moreover, this $R$ only depends on the initialization $w_{\text{init}}$ and the reparameterization $g$, and is independent of the loss function $L_t$.
\end{theorem}

We have used Theorem \ref{Regularization Parameterization : Thm 4.9} to show the Theorem \ref{Time dep mirror : main theorem} in the main text. Moreover, we recover the convergence result for Bregman functions in Theorem \ref{thm : convergence Bregman}. The details of these results are presented in Appendix \ref{appendix: main proofs}.

\newpage

\section{Proofs of Section \ref{section : theory}}\label{appendix: main proofs}

\begin{theorem}\label{Time dep mirror : main theorem appendix}
    Let $g : M \rightarrow \mathbb{R}^n$ and $h : M \rightarrow \mathbb{R}$ be regular and commuting parameterizations satisfying Assumption \ref{Regularization Parameterization : Ass 3.5}. Then there exists a time-dependent Legendre function $R_{a}$ such that 
    \begin{equation*}
        d \nabla_x R_{a_t}(x_t) = -\nabla_x f(x_t) dt, \qquad x_0 = g(w_{init})
    \end{equation*}
    where $a_t = -\int_0^t \alpha_s ds$.
    Moreover, $R_a$ only depends on the initialization $w_{\text{init}}$ and the reparameterization $g$ and $h$ and is independent of the loss function $f$.
\end{theorem}
Proof. 
Consider the time-dependent loss function $L_t(x,y) = f(x) + \alpha_t y $. Applying Theorem \ref{Regularization Parameterization : Thm 4.9} implies there is a Legendre function $R(x,y)$ such that
\begin{equation}\label{legendre : start}
    \begin{cases}
        \nabla_x R(x_t,y_t) = - \int_0^t \nabla_x f(x_s) ds \\
        \partial_y R(x_t, y_t) = - \int_0^t \alpha_s ds.
    \end{cases}
\end{equation}
We use Eq.~(\ref{legendre : start}) to derive the time-dependent Legendre function. 
First note that $\partial_y \partial_y R(x, y) > 0$ for $(x,y) \in dom R$ since $R$ is strictly convex. 
This implies that the map $y \rightarrow \partial_y R(x, y)$ is invertible. 
Let the inverse be denoted by $Q(x, a)$, where in the dynamics $a_t = - \int_0^t \alpha_s ds$.
Plugging $Q$ into the first part of Eq.~(\ref{legendre : start}) gives us
\begin{equation}\label{legendre : middle}
    \nabla_x R\left(x_t,Q\left(x_t, a_t\right)\right) = - \int_0^t \nabla_x f\left(x_s\right) ds,
\end{equation}
where $\nabla_x$ is still the derivative with respect to the first entry.
Eq.~(\ref{legendre : middle}) looks already like a time-dependent mirror flow. 
We show now that there exists a Legendre function $R_{\alpha}$ with the map $\nabla_x R\left(x,Q\left(x, \alpha\right)\right)$ as the gradient. 
This we can do by showing that the Hessian is symmetric and positive definite and that the $R_{\alpha}$ is essentially smooth.

By implicitly differentiating, we make the following observation:
\begin{equation*}\label{exp reg : implicit diff}
    \frac{d Q}{dx} = - \frac{1}{\partial_y \partial_y R(x, Q)} \nabla_x \partial_y R(x, Q).
\end{equation*}
Next, we compute the Hessian and apply observation \ref{exp reg : implicit diff}:
\begin{align*}
    \nabla^2_x R_{\alpha} &= \nabla^2_x R(x, Q) + \partial_y \nabla_x R(x,Q) \cdot \frac{d Q}{dx} \\
    & = \nabla^2_x R(x, Q) -\frac{1}{\partial_y \partial_y R(x, Q)} \partial_y \nabla_x R(x,Q)  \nabla_x \partial_y R(x, Q)^T.
\end{align*}
Observe that this matrix is symmetric as it is a sum of symmetric matrices. 
It remains to be shown that the Hessian matrix is positive definite.
For this, we use that $\nabla^2_x R$ and $\partial_y^2 R$ are positive definite. 
This implies that the the Hessian of $R$ inverse is PD. We use the block inversion matrix formula for matrix $M$,
\[
M = \begin{bmatrix}
A & B \\
C & D
\end{bmatrix}
\]
where \( A \) and \( D \) are square and invertible.
For notation clarity, define the Schur complement of \( D \) as:
\[
S = A - B D^{-1} C
\]
Then the inverse of \( M \) is given by:
\[
M^{-1} =
\begin{bmatrix}
S^{-1} & -S^{-1} B D^{-1} \\
- D^{-1} C S^{-1} & D^{-1} + D^{-1} C S^{-1} B D^{-1}
\end{bmatrix}
\]
The first block entry of the matrix $\nabla^2 R$ is given by the inverse Shur complement:
\begin{equation*}
    \left(\nabla^2_x R(x, y) -\frac{1}{\partial_y \partial_y R(x, y)} \partial_y \nabla_x R(x,y)  \nabla_x \partial_y R(x, y)^T\right)^{-1}
\end{equation*}
which is also PD. Now this implies the result as the inverse of $\nabla^2_x R_{\alpha}$ is PD.
It follows that there exists a function $R_{a}$ such that $\nabla R_{a} = \nabla_x R(x,Q(x, a))$ by Corollary 16.27 in \citep{lee2013introduction}, concluding the first part.

Finally, $R_{a}$ is essentially smooth by construction, using that $R$ is essentially smooth. 
The boundary $bn(R_a)$ by construction is the set of points $x^*$ that have a sequence $x_n \in dom int \nabla_x R(\cdot , Q(\cdot,a))$ such that if $x_n \rightarrow x^*$ we have $|||\nabla R|| \rightarrow \infty$.
Suppose that $R_{a}$ is not essentially smooth then there exists a sequence $\{x_n\}$ with $x_n \rightarrow 
bd(R_{a})$ as $n \rightarrow \infty$ such that $\lim_{n \rightarrow \infty} 
||\nabla_x R(x_n, Q(x_n, a))||^2_{L_2} < \infty$.
Nevertheless, $R$ is essentially smooth this implies that 
\begin{equation*}
    \lim_{n \rightarrow \infty} ||\nabla_y R(x_n, Q(x_n, a))||^2 = a^2 = \infty,
\end{equation*}
leading to a contradiction.
Hence, $R_{a}$ is a Legendre function with the domain similarly constructed as the boundary. $\square$

\begin{theorem}\label{thm : convergence Bregman appendix}
    Assume the same settings as Theorem \ref{Time dep mirror : main theorem}. 
    Furthermore assume that for $\alpha_t \geq 0$ there is a $T > 0$ such that for $t \geq T$, $\alpha_t =0$. Moreover for $a \in [b, 0]$, $R_{a}$ is a contracting Bregman function for some $b < 0$. Assume that for all $t \geq 0$ the integral $a_t : = -\int_0^t \alpha_s ds \geq b$. 
    For the loss function assume that $\nabla_x f$ is locally Lipschitz and $\text{argmin} \{ f(x) : x \in \text{dom} R_{a_{\infty}} \} $ is non-empty. Then if $f$ is quasi-convex $x_t$ converges to a point $x_*$ which satisfies $\nabla_x f(x_*)^T \left(x - x_*\right) \geq 0$ for $x \in dom R_{a_{\infty}}$. 
    In addition if $f$ is convex $x_*$ converges to a minimizer $f$ in $\overline{\text{dom} R_{a_{\infty}}}$. 
\end{theorem}
Proof.
We can bound the trajectory of $x_t$ by using the time-dependent Bregman divergence. The divergence between a critical point $x^*$ of $f$ and the iterates $x_t$ is given by
\begin{equation*}
     D_{a_t}(x^*, x_t) := R_{a_t}(x^*) - R_{a_t}(x_t) - \nabla_x R_{a_t}(x_t)^T (x^* - x_t) \geq 0
\end{equation*}

Note that the contracting property implies that for $a_2 \leq a_1$ we have $\text{dom} R_{a_2} \subset \text{dom} R_{a_1} $. Thus, a critical point $x^*$ in $\text{dom} R_{a_{\infty}}$ is in $\text{dom} R_{a_t}$. Hence, the divergence is well-defined. 
Due to that $f$ is quasi convex and $R_a$ contracting we have that $D_{a_t}(x^*, x_t)$ is bounded. 
From the contracting property it follows that $ R_{a_{\infty}}(x^*)  \geq  R_{a_t}(x^*) $.
By definition of a Bregman function, we have that $x_t$ stays bounded for all $t \geq 0$.
It follows that $x_T$ is in the domain of $R_{a_{\infty}}$ and bounded.
Therefore, we have that $D_{a_t}(x^*, x_t) \leq R_{a_{\infty}}(x^*) - R_{a_t}(x_t) - \nabla_x R_{a_t}(x_t)^T (x^* - x_t) =: W_t$.
Now we show that the evolution of $W_t$ is decaying, implying that $D_{a_t}(x^*, x_t)$ is bounded.
The evolution is given by
\begin{align*}
    d W_t &= \alpha_t \frac{d}{da_t}  R_{a_t}(x_t) dt - \nabla_x R_{a_t}(x_t) dx_t +  \nabla_x R_{a_t}(x_t) dx_t - d\nabla_xR_{a_t}(x_t)^T (x^* - x_t) \\
    &\leq -  d\nabla_xR_{a_t}(x_t)^T (x^* - x_t)
    \\ &= + d\nabla_x f(x_t)^T (x^* - x_t) \\ &\leq 0
\end{align*}
where we used that $\alpha_t \geq 0$ and the contracting property for the first inequality, the time dependendent mirror flow relationship in the second and quasi-convexity for the last.
Therefore $x_t$ stays bounded for $t \in [0,T]$.
Now, using the geometeric interpretation Eq.~(\ref{exp reg : geom interpretation}) we have that the evolution of $\tilde{x}_t = x_{T+t}$ is a gradient flow on a Riemannian manifold with metric $\left(\nabla^2_x R_{a_{\infty}}\right)^{-1}$. 
Therefore Theorem 4.14 in \citep{Li2022ImplicitBO} applies, which concludes the result. $\square$

\newpage

\section{Optimality characterizing the implicit bias.}\label{appendix: section opt}
In this section, we state a general result for underdetermined linear regression extending Theorem 4.17 \citep{Li2022ImplicitBO}. Moreover, we state a detailed result for matrix sensing extending Corollary 6 \citep{Wu2021ImplicitRI}.

 For underdetermined linear regression, let $\{(z_i,y_i)\}_{i=1}^n \subset \mathbb{R}^d \times \mathbb{R}$ be a dataset of size $n$.
 Given a reparameterization $g$ with regularization $h$, the output of the linear model on the $i$-th data is $z_i^T g(w)$. The goal is
 to solve the regression for the label vector $Y = (y_1,y_2,\hdots,y_n)^T$. For notational convenience, we
 define $Z = (z_1,z_2,\hdots,z_n) \in \mathbb{R}^{n \times n}$.

In order to show optimality for underdetermined linear regression we use the following Lemma:

\begin{lemma}(Lemma B.1 \citep{Li2022ImplicitBO})\label{lemma : optimality linear}
    For any convex function $R : \mathbb{R}^d \rightarrow R \cup \{\infty\}$ and 
    $Z \in \mathbb{R}^{n\times d}$, suppose 
    $\nabla R(x^*) = Z^T \lambda$ for some 
    $\lambda \in \mathbb{R}^n$, then
    \begin{equation*}
        R(x^*) = \min_{x : Z(x-x^*) = 0}R(x).
    \end{equation*}
\end{lemma}

We now denote the function to be optimized by $f(x) = \tilde{f}(Zx -Y)$, to emphasize the linearity of the optimization problem.

\begin{theorem}\label{appendix : opt thm}
    Assume the same settings as Theorem \ref{Time dep mirror : main theorem}. 
    Furthermore, assume that for $\alpha_t \geq 0$ there is a $T > 0$ such that for $t \geq T$, $\alpha_t =0$. If $x_t$ converges as $t \rightarrow \infty $ and the convergence point $x_{\infty} = \lim_{t \rightarrow \infty} x_t$ satisfies $Z x_{\infty} = Y$, then
    \begin{equation}\label{equation : opt app}
        x_{\infty} = \text{argmin}_{x: Z x = Y} R_{a_T}(x).
    \end{equation}
    Therefore, the gradient flow minimizes the changed regularizer $R_{a_T}$ among all potential solutions. $\square$
\end{theorem}
Proof. 
Since we assume convergence i.e. $Zw_{\infty} = y$, we have to show the KKT condition associated with Eq.~(\ref{equation : opt app}) is satisfied. 
We have to show that $\nabla R_{a_T}(x^*) \in \text{span}\left(Z^T\right)$.
This follows directly from integrating the time-dependent mirror flow for $t \geq T$:
\begin{equation*}
    \nabla R_{a_T}(x_t) - \nabla R_{a_0}(x_0) = - Z^T \int_0^t \nabla \tilde{f}\left(Z x_s - Y\right)ds \in \text{span}\left( Z^T \right).
\end{equation*}
Notice by definition of the Legendre function and its convex conjugate we have that $\nabla R_{a_0}(x_0)  = 0$.
Therefore, $\nabla R_{a_T}(x_t) \in \text{span}\left(Z^T\right)$, which further implies that $\nabla R_{a_T}(x_{\infty}) \in \text{span}\left(Z^T\right)$. Applying Lemma \ref{lemma : optimality linear} concludes the result. $\square$

Theorem \ref{appendix : opt thm} shows optimality for underdetermined linear regression. Moreover, together with Theorem \ref{thm : quadratic param} we extend results by a series of papers on quadratic reparameterizations \citep{jacobs2024maskmirrorimplicitsparsification, Li2022ImplicitBO, gunasekar2017implicit, azulay2021implicitbiasinitializationshape, Wu2021ImplicitRI}.

We will focus now focus on one particular example: matrix sensing \citep{Wu2021ImplicitRI, gunasekar2017implicit}. 
The reason for this focus is to show the effect on the spectrum on the matrix.
We show we can modulate between the Frobenius norm and nuclear norm similar to the modulation between $L_2$ and $L_1$ regularization as in \citep{jacobs2024maskmirrorimplicitsparsification}.
Moreover, we can induce a new grokking effect distinct from \citep{lyu2024dichotomyearlylatephase,Liu2022OmnigrokGB}, which considers large initialization and small weight decay.

Denote by $A_i \in \mathbb{R}^{n \times n}$ with $i \in [m]$ the sensing matrices and consider the loss function $f(X) = \frac{1}{2m } \sum_{i=1}^m \left( \langle A_i, X\rangle -y_i\right)^2$. Moreover, let $\mathbb{S}_n^+$ be the class of symmetric positive semi-definite matrices of size $n \times n$. 

\begin{corollary}\label{corrolary : opt sensing}
Assume that the sensing matrices $A_i$’s are symmetric and commute, and that
there exists a $X^* \in \mathbb{S}_n^+$ satisfying $f(X^*) = 0$. Moreover, assume that for $\alpha_t \geq 0$ there is a $T > 0$ such that for $t \geq T$, $\alpha_t =0$.  Then, the gradient flow defined by $\frac{dU_t}{
 dt} =-\nabla_X f(U_t U_t^T) U_t - \alpha_t U_t$
 and any initialization satisfying $U_0U^T_0 = \beta I$ converges to a matrix $U_{\infty}$ minimizing
 \begin{equation}\label{eq: eigen opt}
     \sum_{i=1}^n \left( \text{log}\left(\frac{1}{A_T}\right) - 1\right) \lambda_i + \lambda_i \text{log} \lambda_i
 \end{equation}
 among all global minima of f, where $\{\lambda_i\}^n_{i=1}$ denote the eigenvalues of the matrix $X_{\infty} := U_{\infty}U^T_{\infty}$ and $A_T := \beta \exp{\left(-2\int_0^T \alpha_s ds\right)}$.
\end{corollary}
Proof. 
Convergence follows from Theorem \ref{thm : quadratic param} and optimality follows from Theorem \ref{appendix : opt thm}. It is left to show that the corresponding time-dependent Bregman function $R_{a_t}(X_t)$ is given by Eq.~(\ref{eq: eigen opt}). 
From the gradient flow we can derive the time-dependent Bregman function:
\begin{equation*}
    R_{a_t}(X_t) = \text{Tr}\left(X_t \left( \text{log}\left(\frac{1}{\beta \exp{\left(-2\int_0^t \alpha_s ds\right)}}\right) - 1\right) + X_t \text{log} X_t\right).
\end{equation*}
Next, from the eigenvalue decomposition for symmetric matrices and the fact that all the above matrices are simultaneously diagonalizable, we get Eq.~(\ref{eq: eigen opt}). $\square$

\paragraph{Experiment}
To illustrate the implication of Corollary \ref{corrolary : opt sensing} we conduct an experiment on matrix sensing.
The implication is that when we train with weight decay it is stored in the the time-dependent Bregman function.
We can leverage this lasting effect by turning off the weight decay and reach the sparse (optimal) solution as in Eq.~(\ref{equation : opt app}).
This allows us to induce a grokking-like phenomenon.

We use a similar experimental setup as in \citep{Wu2021ImplicitRI}. 
Specifically, we generate a sparse groundtruth matrix $X^*$ by first sampling $U^* \in \mathbb{R}^{n\times r}$, where $r =5$, with entries drawn i.i.d. from $N(0,1)$. We then set $X^* = U^* \left(U^*\right)^T$ and normalize it such that $||X^*||_{nuc} = 1$. We generate $m$ symmetric sensing matrices $A_i := \frac{1}{2}\left( B_i + B_i^T\right) $, where the $B_i$ entries are drawn i.i.d. from $N(0,1)$. We use learning rate $\eta = 0.25$, initialize $U_0 = I \beta$ with $\beta =0.1$ and train for $5000$ steps. We consider $3$ scenarios:
\begin{itemize}
    \item Train without weight decay i.e. $\alpha =0$
    \item Train with weight decay $\alpha = 0.01, 0.02$
    \item Train with weight decay $\alpha = 0.02$ for $2500$ steps and after that turn weight decay off, i.e. $\alpha = 0$.
\end{itemize}
The second scenario ($\alpha = 0.01$) and the third scenario are constructed such that the same amount of total regularization is applied at the end of training.
Moreover, the same setup is used for the linear paramterization with $L_1$ regularization.

\begin{figure}[!htb]
  \centering
    \begin{subfigure}[b]{0.45\textwidth}
    \centering
    \includegraphics[width=1.0\textwidth]{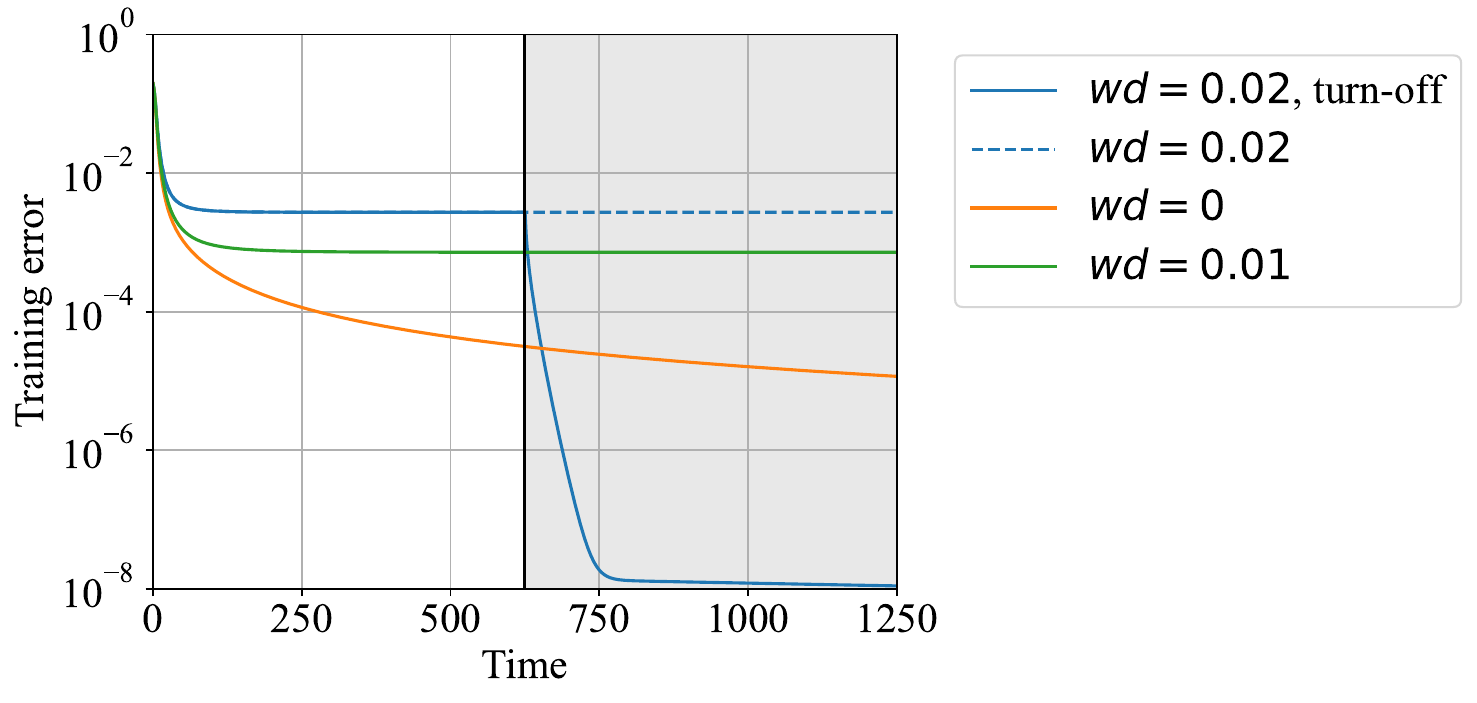}
    \caption{Training error.}
    \label{fig: sensing training}
    \end{subfigure}
    \hfill
    \begin{subfigure}[b]{0.45\textwidth}
    \centering
    \includegraphics[width = 1.0 \textwidth]{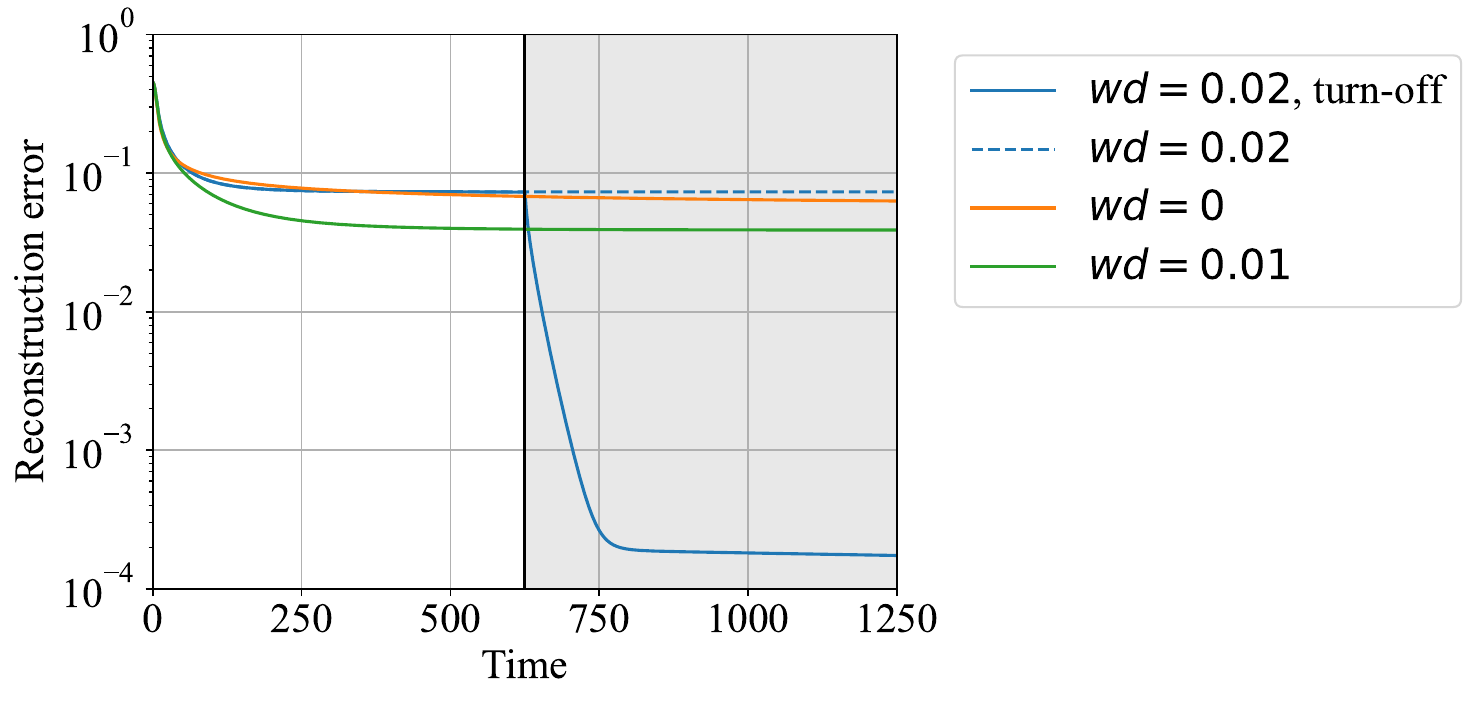}
    \caption{Reconstruction error.}
    \label{fig: sensing reconstruction}
    \end{subfigure}
    \caption{Train and reconstruction error for the matrix sensing experiment for quadratic parameterizations. Observe that both the training and reconstruction error decrease when the weight decay is turned off recovering the sparse ground truth.}
    \label{fig: sensing}
\end{figure}

\begin{figure}[!htb]
  \centering
    \begin{subfigure}[b]{0.45\textwidth}
    \centering
    \includegraphics[width=1.0\textwidth]{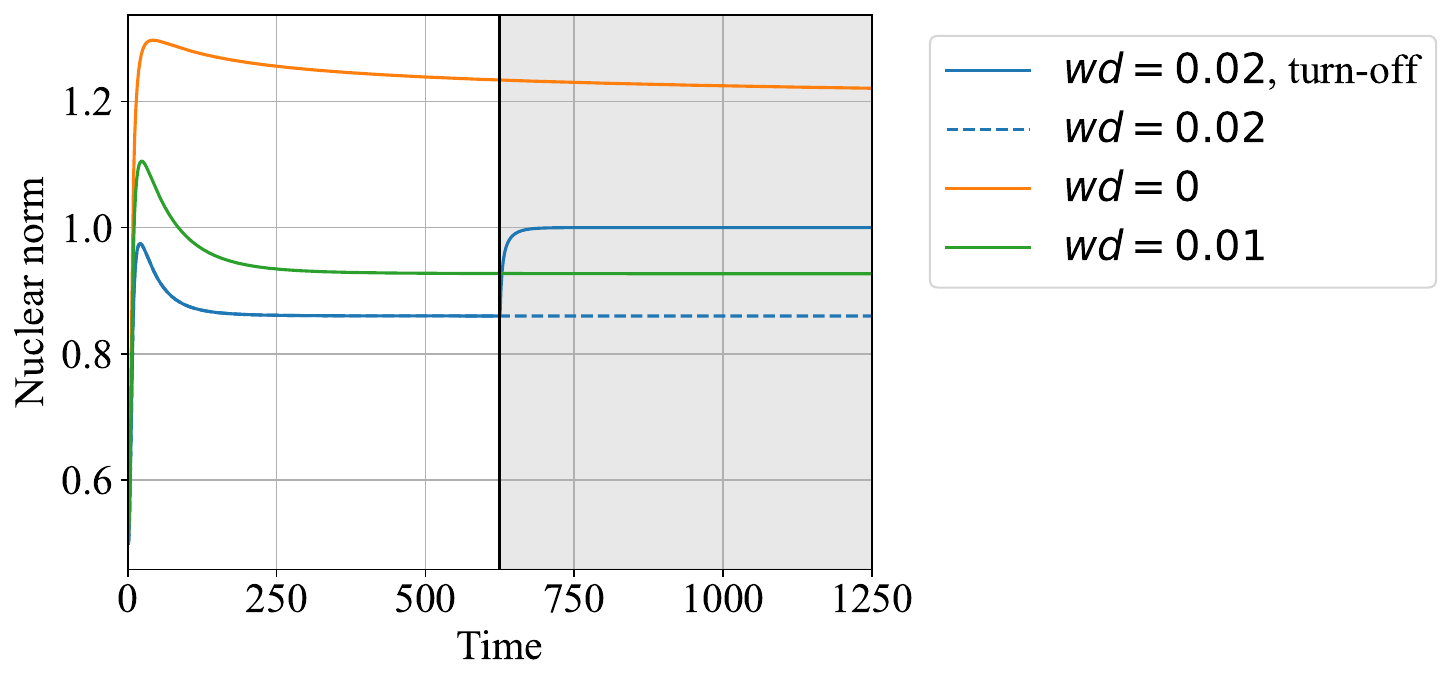}
    \caption{Quadratic reparameterization.}
    \label{fig: sensing nuc quad}
    \end{subfigure}
    \hfill
    \begin{subfigure}[b]{0.45\textwidth}
    \centering
    \includegraphics[width = 1.0 \textwidth]{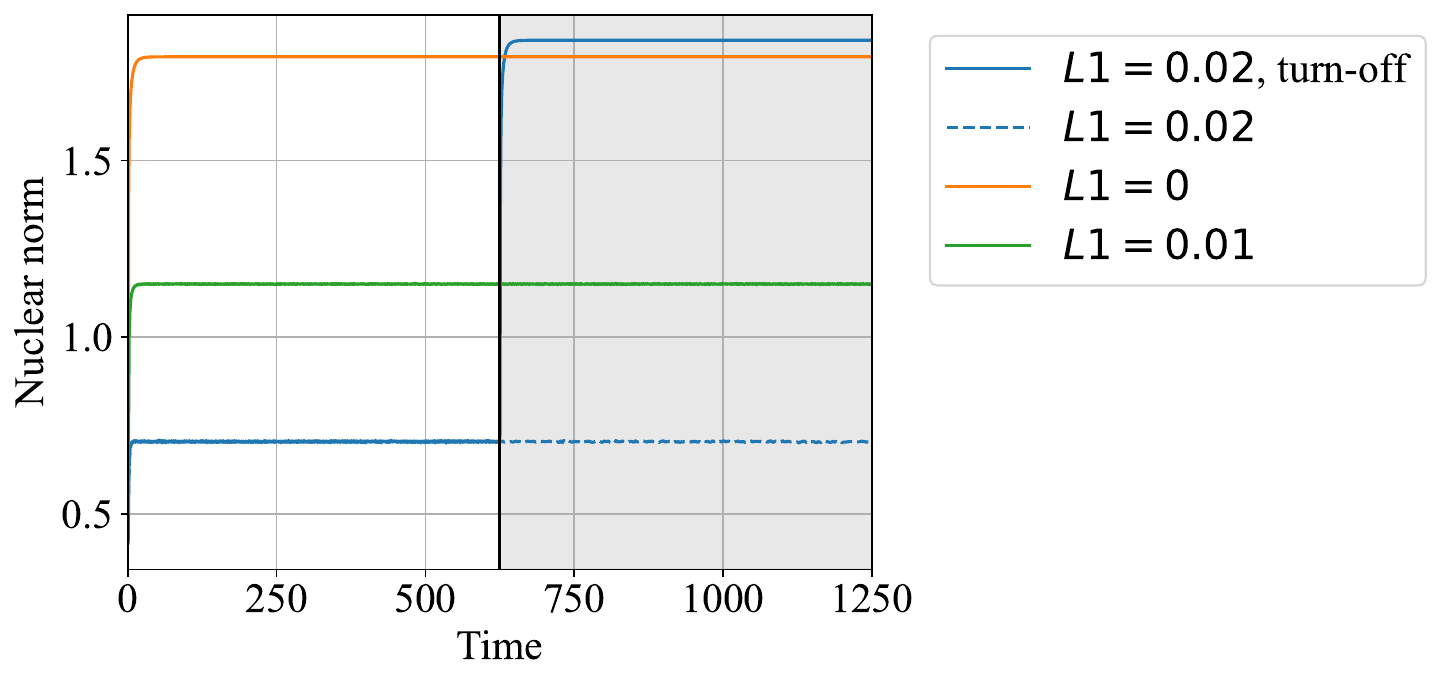}
    \caption{Linear reparameterization.}
    \label{fig: sensing nuc lin}
    \end{subfigure}
    \caption{Tracking the nuclear norm for both quadratic and linear reparameterization. In the case of the quadratic parameterization, the effect of the regularization gets stored resulting in a better reconstruction error in Figure. \ref{fig:sensing R}, whereas this does not happen in the linear case. In the linear case, the nuclear norm increases to the level of that without explicit $L_1$-regularization.}
    \label{fig: sensing nuc}
\end{figure}

In Figure. \ref{fig: sensing} we present the evolution of the training and reconstruction error $||X^* - X_t||_{fro}^2$. We observe that in all scenarios the training error is below $10^{-2}$. In contrast, only the reconstruction error for the third scenario where we turn off the weight decay goes below $10^{-2}$. 
Therefore, by accessing the stored weight decay we can reach closer to the sparse solution illustrating Corollary \ref{corrolary : opt sensing}. 
This is also confirmed by the evolution of the nuclear norm in Figure. \ref{fig: sensing nuc}.

Note that keeping the weight decay on in the second scenario ($wd =0.1$) also leads to a better reconstruction error than the overfitting of scenario one ($wd =0$). 
Nevertheless, due to the explicit trade-off optimality is not possible.
This is not in contradiction with the analysis in \citep{lyu2024dichotomyearlylatephase}, where very small weight decay is used.
In the case of small weight decay, the trade-off is negligible.

\begin{figure}[!htb]
    \centering
    \includegraphics[width=0.5\linewidth]{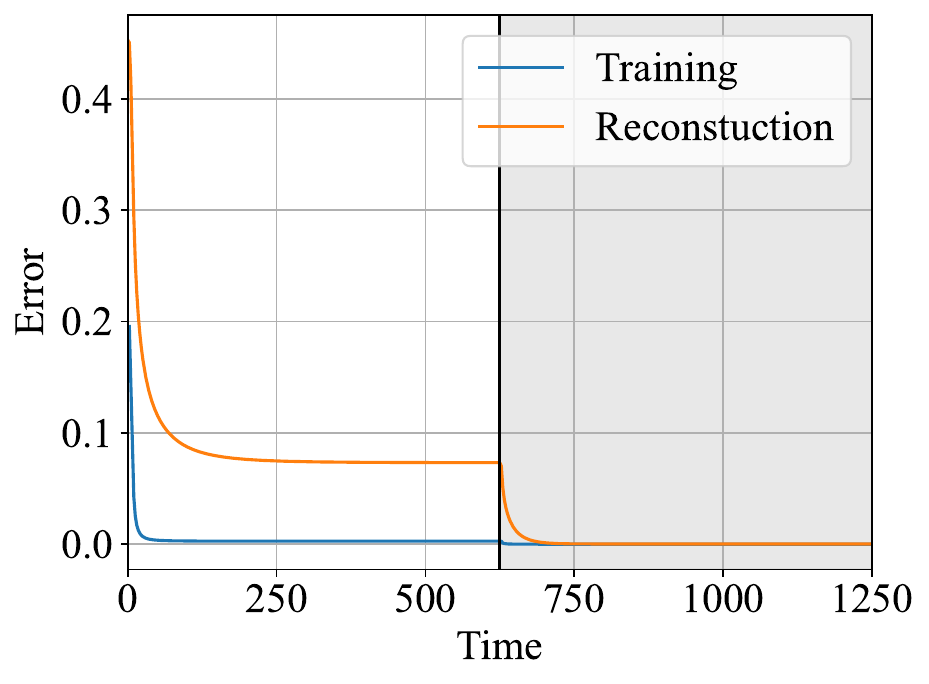}
    \caption{Training and reconstruction error for $\alpha = 0.02$ and turn-off. We recover the sparse ground truth by turning weight decay off for matrix sensing.}
    \label{fig:grok}
\end{figure}

The dynamics in Figure. \ref{fig:grok} are similar to the dynamics of the grokking phenomenon: generalization happens later after no progress. 
Nevertheless, a key difference is that we induce it by turning off the weight decay in contrast to \citep{Liu2022OmnigrokGB, lyu2024dichotomyearlylatephase}. 
This leads to relatively "fast-grokking".

\paragraph{Ablation with different schedules}
Consider the family of schedules with constant regularization strength up to a specific time $T_i$ such that $\alpha_t = \alpha_0$ for $t \leq T_i$ and $\alpha_t = 0$ afterward, for $\alpha_0$ a constant. We choose $\alpha_0 = 0.02$ and $T_i = T/2$. In addition, we consider a linear and cosine decay schedule for the regularization with the same total strength (i.e., same integral), but the regularization is switched off after half of the training time to ensure convergence. 
To compare with the effect of turning off (t-o) the regularization, we include the constant schedule with regularization strength $\alpha_0 = 0.01$.

We observe in Table \ref{tab : ablation sensing} that all schedules with decay or turn-off (t-o) converge to a solution with the same nuclear norm of the ground truth, confirming Theorem 3.6, while the constant schedule does not reach the ground truth. 

\begin{table}[h!]
\centering
\caption{Performance of different regularization schedules.}\label{tab : ablation sensing}
\begin{tabular}{@{}l|cccc@{}}

Schedule & Nuclear norm & Train loss & Rec error & Time to $10^{-7}$ train loss \\
\midrule
Constant 0.01, no t-o   & 0.93 & $7.2 \times 10^{-4}$ & $3.9 \times 10^{-2}$ & -   \\
Linear decay            & 1.00 & $1.8 \times 10^{-8}$ & $2.3 \times 10^{-4}$ & 661 \\
Cosine decay            & 1.00 & $1.7 \times 10^{-8}$ & $2.1 \times 10^{-4}$ & 624 \\
Constant 0.02, t-o      & 1.00 & $1.1 \times 10^{-8}$ & $1.7 \times 10^{-4}$ & 716 \\
Constant 0.2, t-o       & 1.00 & $2.7 \times 10^{-10}$ & $2.7 \times 10^{-5}$ & 209 \\
Constant 2, t-o         & 1.00 & $2.1 \times 10^{-10}$ & $2.4 \times 10^{-5}$ & 209 \\
Constant 20, t-o        & 1.00 & $7.9 \times 10^{-13}$ & $1.4 \times 10^{-6}$ & 239 \\

\end{tabular}

\end{table}

\newpage

\section{Deeper reparameterization}\label{appendix : other param}
In this section, we explore several reparameterizations and limitations of the framework. 
We show that Theorem \ref{Time dep mirror : main theorem} does not apply to linear parametrization. Moreover, Theorem \ref{Time dep mirror : main theorem} does not apply to overparameterizations with a depth larger than $2$ and weight decay. Nevertheless, we show in experiments that similar effects can occur. 
We illustrate both the type change and range shrinking effect.

\paragraph{Linear parametrization}
From Corollary \ref{corollary : seperable}, we derive another corollary for non-overparameterized parametrization.
\begin{corollary}\label{normal param}
    Let $g(x) = x$ be the identity parametrization and $h \in C^2(\mathbb{R}^n, \mathbb{R})$. Then Theorem \ref{Time dep mirror : main theorem} applies if and only if, $h$ is given by
$
        h(x) = \sum_{i = 1}^n c_i x_i + d
$
    where $c_i, d \in \mathbb{R}$ are arbitrary coefficients.
\end{corollary}
Proof. To apply the theorem, $h$ needs to be commuting with $g$, implying that $
    \partial_i \partial_i h = 0$ $\forall i \in [n]$, 
concluding the result. $\square$

Corollary \ref{normal param} poses a limitation in the applicability of Theorem \ref{Time dep mirror : main theorem}.
Since $h$ is not positive for all $x \in \mathbb{R}^n$, the resulting optimization problem is ill-posed.
Therefore, standard non-reparameterized loss functions cannot be analyzed in this manner.

\paragraph{Beyond quadratic reparameterization}
We show that the current framework excludes higher-order reparameterization with weight decay. 
In order to show that 

\begin{theorem}\label{thm : high order}
    Let $g : \mathbb{R}^k \rightarrow \mathbb{R}$ be given by $g(w) := \Pi_{i=1}^k w_i$ , a $k > 2$ depth reparamterization. Moreover, let $h : \mathbb{R}^k\rightarrow \mathbb{R}$ and $h(w) = \sum_{i=1}^k w_i^2$. 
    Then $g$ and $h$ do not commute.
\end{theorem}
Proof. This follows directly from checking the commuting condition between $g$ and $h$:
\begin{align*}
    [\nabla_w g, \nabla_w h](w) &= \nabla_w g(w) \nabla^2_w h(w) - \nabla_w h(w) \nabla^2_w g(w) \\ &= \begin{bmatrix}
        \left(4 - 2k\right) \Pi_{i \in [k]\setminus\{1\}} w_i \\
        \vdots \\
        \left(4 - 2k\right) \Pi_{i \in [k]\setminus\{k\}} w_i
    \end{bmatrix}.
\end{align*}
In order for this to be equal to zero all products need to be zero.
This implies that the gradient flow given by
\begin{equation*}
    dw_t = - \begin{bmatrix}
        \Pi_{i \in [k]\setminus\{1\}} w_{i,t} \\
        \vdots \\
        \Pi_{i \in [k]\setminus\{k\}} w_{i,t}
    \end{bmatrix} \odot \nabla_x f(g(w_t)) - \alpha_t w_tdt,
\end{equation*}
becomes $dw_t = - \alpha_t w_t dt$ and is independent of $f$.
Hence, $g$ and $h$ do not commute 
$\square$

Theorem \ref{thm : high order} implies that we can not apply Theorem \ref{Time dep mirror : main theorem}. We note that the commuting condition is only a sufficient criterion such that a pair $(g,h)$ is a time-dependent mirror flow.

\paragraph{Experiment}
Although our theoretical result does not hold for reparametrizations with higher depth, we illustrate that the expected effects do occur as well for higher depth. 
We consider the reparameterization $m \odot w \odot v$ for diagonal linear networks and compare with $m \odot w$, both with weight decay. Moreover, we compare with the reparameterization $m$ with $L_1$ regularization to motivate the importance of the geometry, which is controlled by the time-dependent Legendre function.
This is similar to the matrix sensing case explored in the previous section and main paper.

Let $d = 40$ be the amount of data points and $n = 100$ the dimension of the data.
We generate independent data $Z_k \sim N(0, I_n)$ for $k \in [d]$.
We assume a sparse ground truth $x^*$ such that $||x^*||_{L_0} = 5$.
The training labels are generated by $y_k = Z_k^T x^*$.
Moreover, the mean squared error loss function is used.
The learning rate $\eta = 10^{-3}$ and we use weight decay $\alpha \in \{0.01, 0.1, 1\} $.
We run the $100000$ steps with weight decay, after that we run the same amount of steps without weight decay. 
We initialize $m = \textbf{0}$ and $w = z = \textbf{1}$, this ensures that both parametrizations are initialized at zero and have the same scaling.
In this setup, we illustrate the type change similarly predicted for the parametrization $m \odot w$.
Moreover, we illustrate the range shrinking which occurs for higher depth parametrization $u^{2k} - v^{2k}$. Note that the ground truth has the following ratio between the $L_1$ and $L_2$ norm $2.23$.

\begin{figure}[ht]
\centering
\begin{subfigure}{0.3\textwidth}
         \includegraphics[width=1\textwidth]{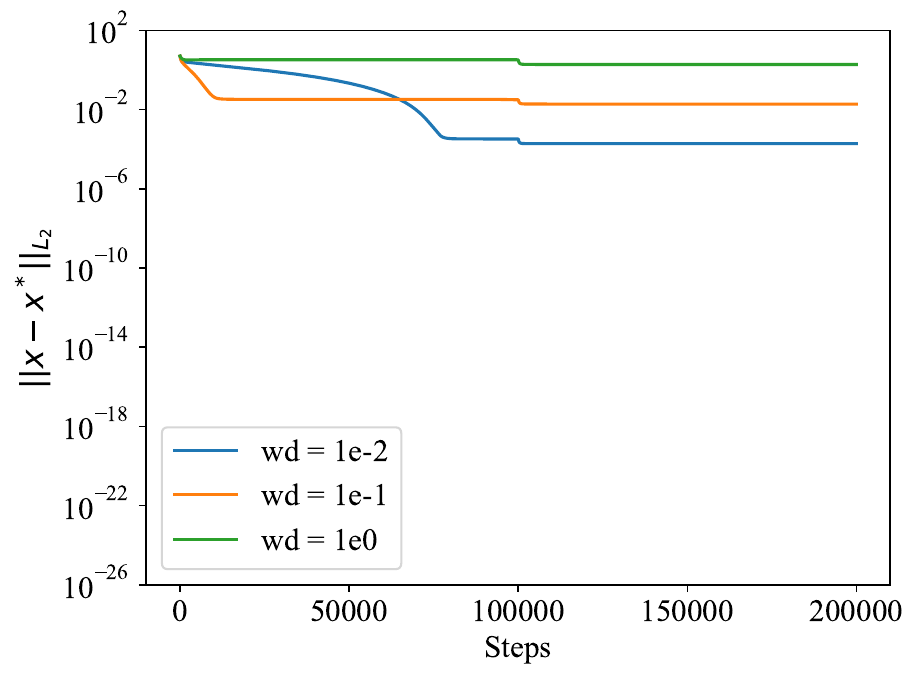}
        \caption{The evolution of $m$.}
       \label{fig: evol m}
\end{subfigure}
\hfill
\begin{subfigure}{0.3\textwidth}
         \includegraphics[width=1\textwidth]{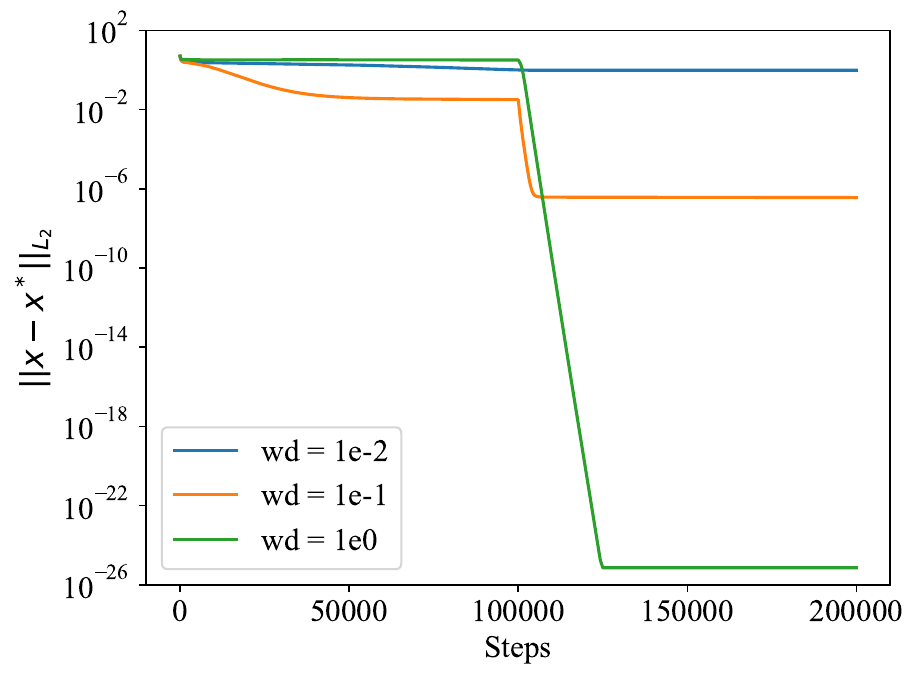}
        \caption{The evolution of $m \odot w$.}
       \label{fig: evol mw}
\end{subfigure}
\hfill
\begin{subfigure}{0.3\textwidth}
         \includegraphics[width=1\textwidth]{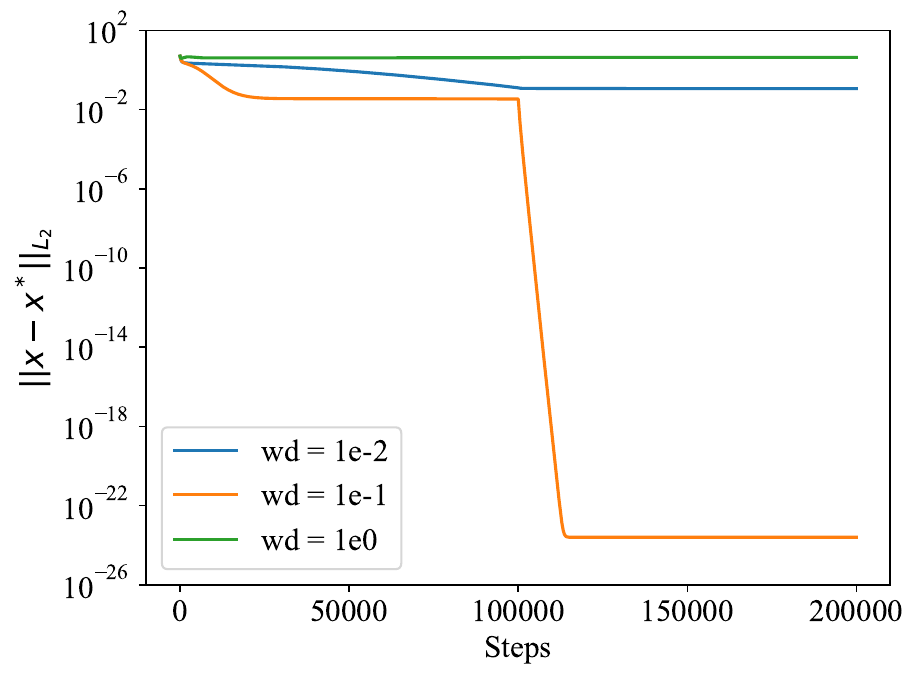}
        \caption{The evolution of $m \odot w \odot z$.}
       \label{fig: evol mwz}
\end{subfigure}
\caption{Illustration of the effect of weight decay with higher order reparameterizations on generalization performance.}
\end{figure}

\begin{figure}[ht]
\centering
\begin{subfigure}{0.3\textwidth}
         \includegraphics[width=1\textwidth]{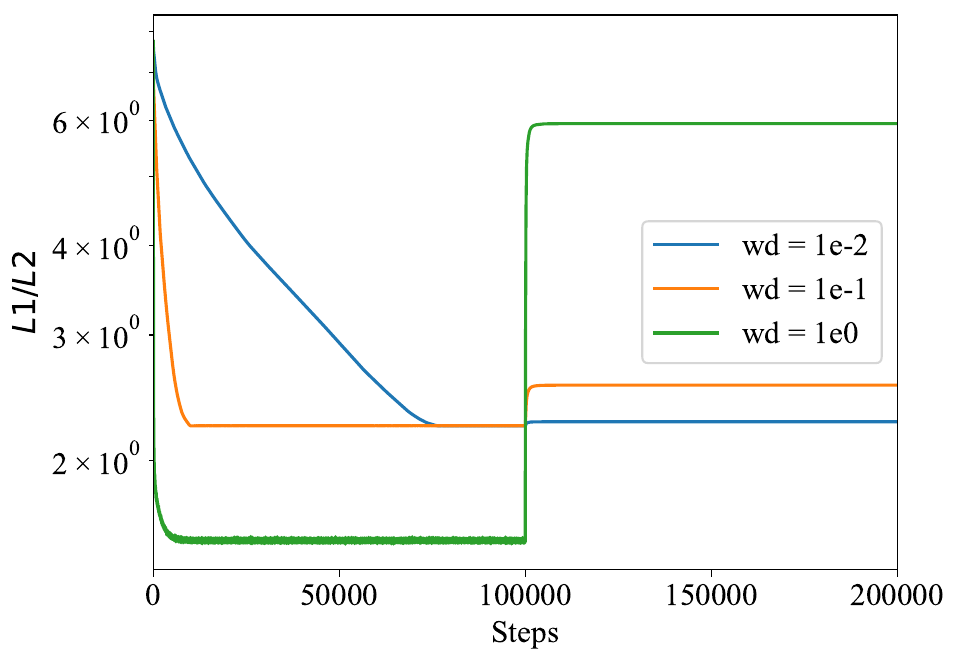}
        \caption{The evolution of $m$.}
       \label{fig: evol m ratio}
\end{subfigure}
\hfill
\begin{subfigure}{0.3\textwidth}
         \includegraphics[width=1\textwidth]{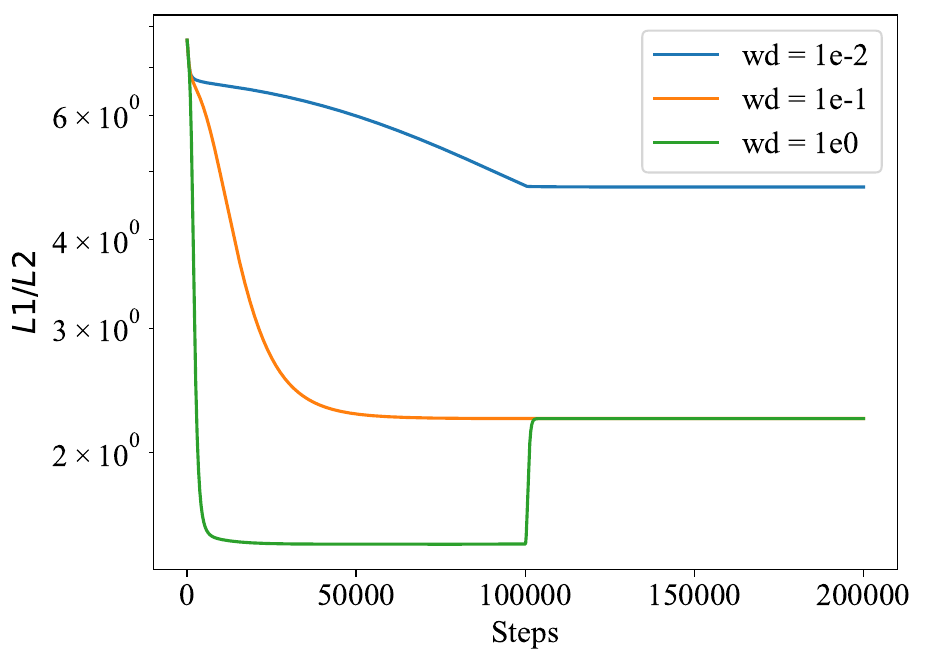}
        \caption{The evolution of $m \odot w$.}
       \label{fig: evol mw ratio}
\end{subfigure}
\hfill
\begin{subfigure}{0.3\textwidth}
         \includegraphics[width=1\textwidth]{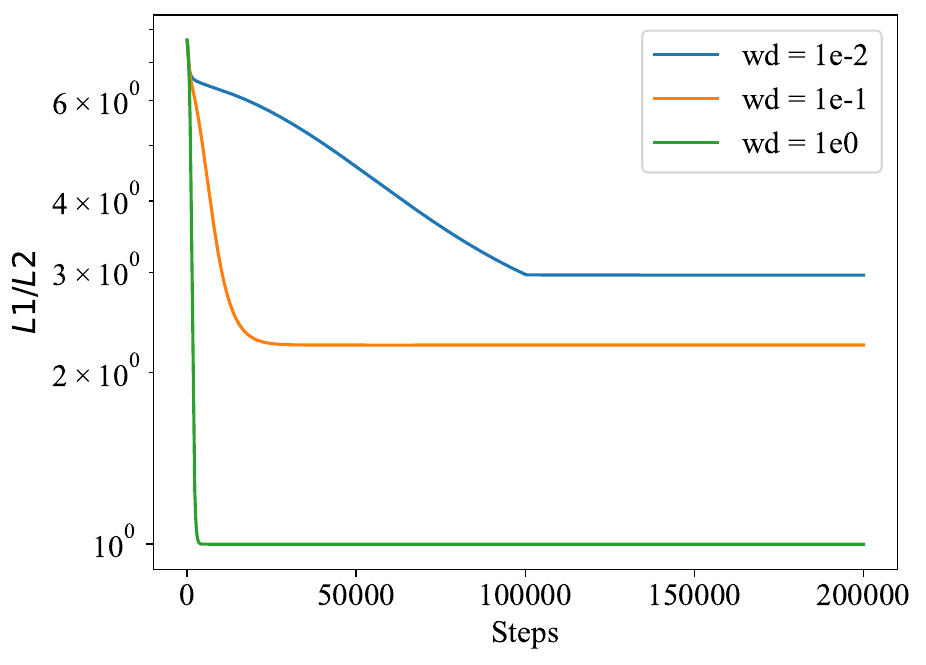}
        \caption{The evolution of $m \odot w \odot z$.}
       \label{fig: evol mwz ratio}
\end{subfigure}
\caption{The ratio between the $L_1$ and $L_2$ for diagonal linear networks.}
\end{figure}

In Figure.\ref{fig: evol m} we observe for $m$ that higher weight decay does not get closer to the ground truth after turning the $L_1$ regularization off.
This is in line with the fact that the regularization is not stored in the geometry as described by Eq.~(\ref{exp reg : geom interpretation}). By turning off the regularization we converge to the closest solution in $L_2$ norm. This is best seen in Figure.\ref{fig: evol m ratio}, where the ratio increases above the value of the ground truth.

In Figure.\ref{fig: evol mw} we observe for $m \odot w$ that higher weight decay gets closer to the ground truth after turning the weight decay off.
This is in line with the fact that the regularization is stored in the geometry as described by Eq.~(\ref{exp reg : geom interpretation}) and a type of bias change from $L_2$ to $L_1$. Furthermore, this is also confirmed in Figure.\ref{fig: evol mw ratio} that for large weight decay, the ratio gets close to the ratio of the ground truth only after turning the weight decay off.
This also illustrates Theorem \ref{thm : convergence Bregman} and \ref{main text : opt thm}.

In Figure.\ref{fig: evol mwz}, we observe for the regularization strength $1e-1$ a similar effect corresponding to the type of bias change from $L_2$ to $L_1$.
In contrast, the higher regularization does not exhibit the same behaviour.
We claim this is due to the range-shrinking effect.
To motivate this is not due to the dynamics getting stuck at $x = \textbf{0}$ we report the final value of the first parameter. The value is equal to $1.58$ which is not equal to either $0$ or the ground truth value $1$. To add to this, in Figure.\ref{fig: evol mwz ratio} we unveil that the ratio for large weight decay stays constant.

In conclusion, the type of bias can improve generalization, whereas $m \odot w$ even goes to the ground truth with high regularization, $m$ does not. Moreover, when we use higher order reparametrization such as $m \odot w \odot z$ we encounter a different phenomenon: range shrinking. To add to this, higher-order parametrization still exhibits the type of bias change in a certain range of regularization strength.
Thus, our theoretical framework leads to verifiable predictions. These can be used to improve the training dynamics of neural networks in general.

\newpage 

\section{Sparse coding}\label{appendix exp log}
We extend our study to the traditional sparse coding problem, with our proposed reparameterization substituting the standard sparse coding step. 
For this experiment, we use the Olivetti faces dataset.
We denote the dictionary with $D$, labels with $z$, the code with $g(u,v)$ and regularization with $h(u,v)$. The feature dimension of $D$ is $n$. This is solved as a linear regression problem with mean squared error. We have used a learning rate $\eta = 0.001/ Lip(D)$ where $Lip(D)$ denotes the resulting Lipschitz constant of the optimization problem depending on the dictionary $D$. 
In addition, we set the number of features $n = 50$ and run for $100$ iterations.

\paragraph{The reparameterization $u^{2k} -v^{2k}$}

\begin{figure}[ht]
         \centering
         \includegraphics[width=0.35\textwidth]{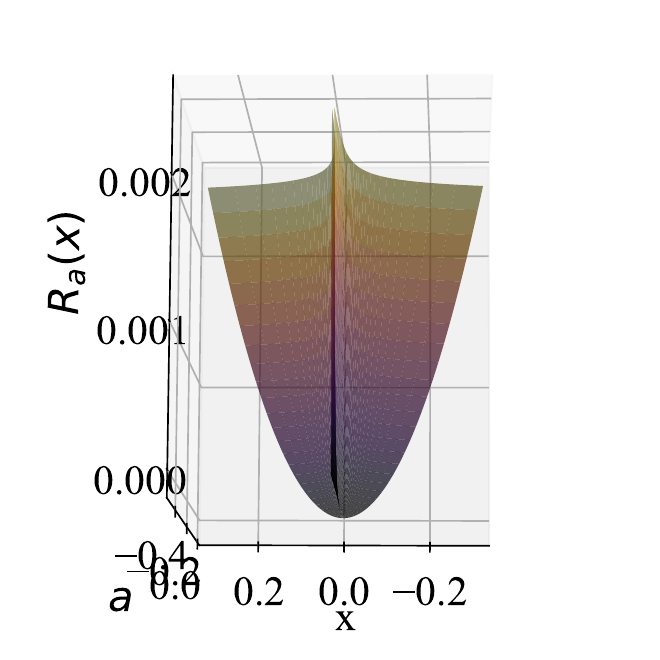}
        \caption{The evolution of the approximated $R_a$ associated with $u^4 -v^4$, where $a = - \int_0^t \alpha_s ds$.}
       \label{fig: Breg evol unvn}
\end{figure}

In this context, we reparameterize the sparse code as $g(u,v)=u^{2k} - v^{2k}$ and set the regularization $h(u,v) =\sum_{i = 1}^n u_i^{2k} + v_i^{2k}$ as discussed in the main paper. 
This parameterization exhibits range shrinking as illustrated by the time-dependent Legendre function in Figure \ref{fig: Breg evol unvn}. 

\begin{figure}[!htb]
  \centering
  \begin{subfigure}{.5\textwidth}
    \centering
    \includegraphics[width=1.0\linewidth]{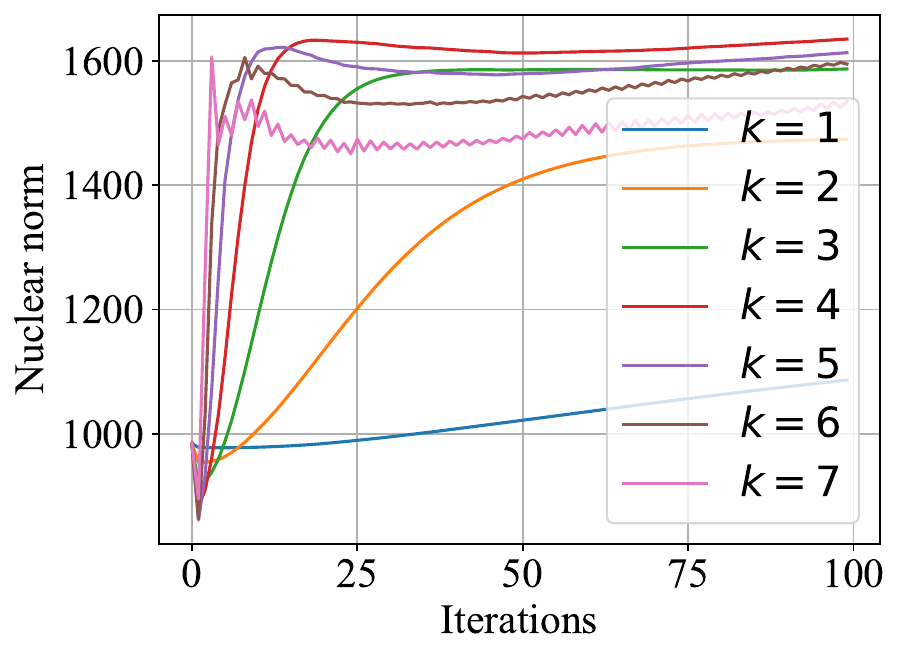}
    \caption{Nuclear norm of sparse code $w$.}
    \label{fig:nuc_norm power}
  \end{subfigure}%
  \begin{subfigure}{.5\textwidth}
    \centering
    \includegraphics[width=1.0\linewidth]{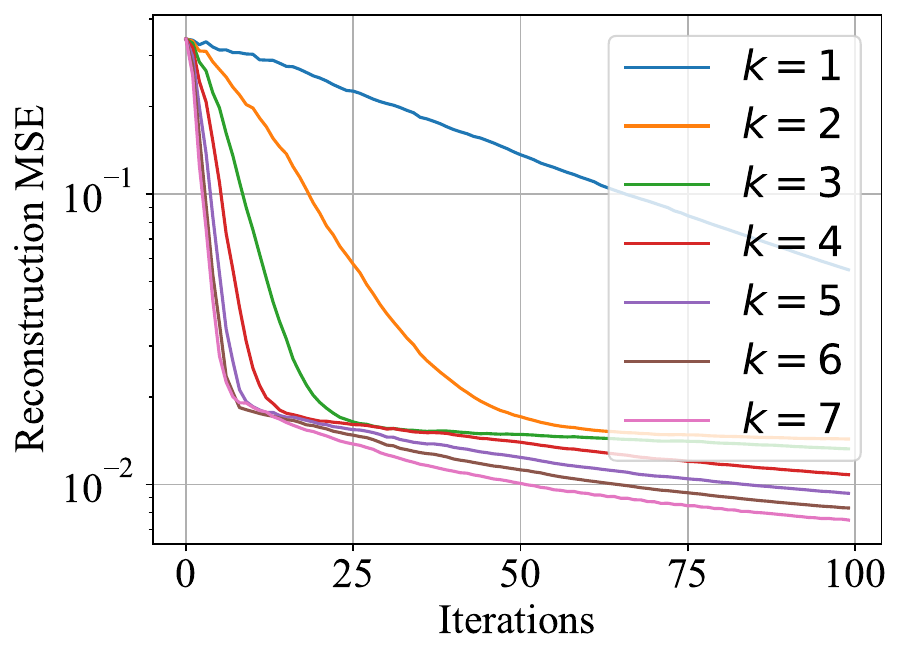}
    \caption{Reconstruction MSE of $x$.}
    \label{fig:subfig2 uk}
  \end{subfigure}
  \caption{Results for sparse coding reparameterisation $g(w)= u^{2k} - v^{2k}$}
  \label{fig:sparse code uk -vk}
\end{figure}

The parameters are initialized as $u_0 = {\frac{1}{2}(\sqrt{x^2+\beta^2}+x)}^{\frac{1}{2k}}$ and $v_0 ={\frac{1}{2}(\sqrt{x^2+\beta^2}-x)}^{\frac{1}{2k}}$, where $\beta=1$, $x\sim\mathcal{N}(0,I_n)$, and all operations are pointwise. We set the regularization strength to $\alpha=0.001$ and explore various values of $k \in [7]$. Throughout the training process, we track two key metrics: the reconstruction mean squared error (MSE) and the nuclear norm of the sparse code, $g(u,v) = u^{2k} - v^{2k}$. The evolution of the nuclear norm is presented in Figure.\ref{fig:nuc_norm power}.
We observe the effect of the range shrinking for $k >1$, for larger $k$ the evolution of the nuclear norm becomes stationary faster.
This indicates that the range in which the time-dependent Legendre function is allowed to move has shrunk.
The shrinking also causes the MSE to converge faster for large $k$, shown in Figure.\ref{fig:subfig2 uk}. 
    

\paragraph{The reparameterization $\log (u) - \log (v)$}\label{appendix log}
We consider a novel reparameterization. 
In the main text, we have seen that the regularization changed the type of bias from $L_2$ to $L_1$.
We now consider a reparameterization with explicit regularization that leads to the opposite type of bias change.
The reparameterization is $ g(w) = \log (u) - \log (v)$.
The regularization found in Corollary \ref{corollary : seperable} is $h(w) = {\sum_{i=1}^n}\log (u_i) + \log (v_i)$.
Then for $u, v > 1$ we can apply Theorem \ref{Time dep mirror : main theorem}.

\begin{figure}[ht]
         \centering
         \includegraphics[width=0.35\textwidth]{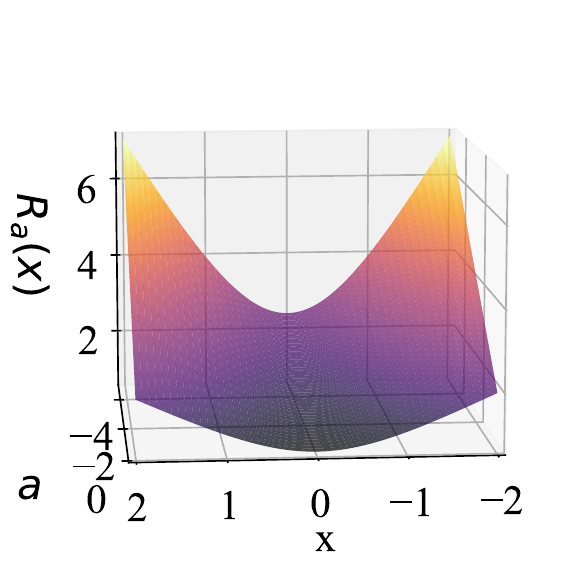}
        \caption{From $L_1$ to $L_2$ implicit bias, with $a = - \int_0^t \alpha_s ds$.}
       \label{fig: Breg evol}
\end{figure}

We now give the resulting time-dependent Legendre function.
The time-dependent Legendre function is
\begin{equation*}
    R_a(x) = \frac{1}{4} \sum_{i = 1}^n \left(u_{0,i}^2-2a\right) \log  \left(e^{-2x_i} + 1\right) + \left(v_{0,i}^2 -2a\right) \log 
    \left(e^{2x_i} + 1\right) \text{ } \forall a < \frac{1}{2}\min \{u_{0,i}^2,v_{0,i}^2\} .
\end{equation*}
The global minimum is centered at $\nabla_x R_a = 0$ and is given by $\log  \left( \sqrt{u_0^2 -2a}\right) -  \log  \left(\sqrt{v_0^2 -2a}\right)$.
Thus a shift occurs when $a$ changes, illustrating the positional bias.
Moreover, to illustrate the type change, consider the balanced initialization $u_0 = v_0 = \beta I$, the Legendre function is then given by
\begin{equation*}
    R_a(x) = \frac{1}{4} \left(\beta^2 -2a\right) \sum_{i = 1}^n \log\left(2 \cosh(x_i)
    \right)
\end{equation*}
which resembles the log-cosh loss function with vertical rescaling. 
The rescaling changes the type of bias from $L_1 \rightarrow L_2$.
The type here is $L_2$ close to the origin and $L_1$ further away from zero.
Due to the scaling, it becomes closer and closer to $L_2$. 
This is illustrated in Figure.\ref{fig: Breg evol}.
Furthermore, we will show in experiments that the type change is crucial for generalization.

\begin{figure}[!htb]
  \centering
  \begin{subfigure}{.5\textwidth}
    \centering
    \includegraphics[width=1.0\linewidth]{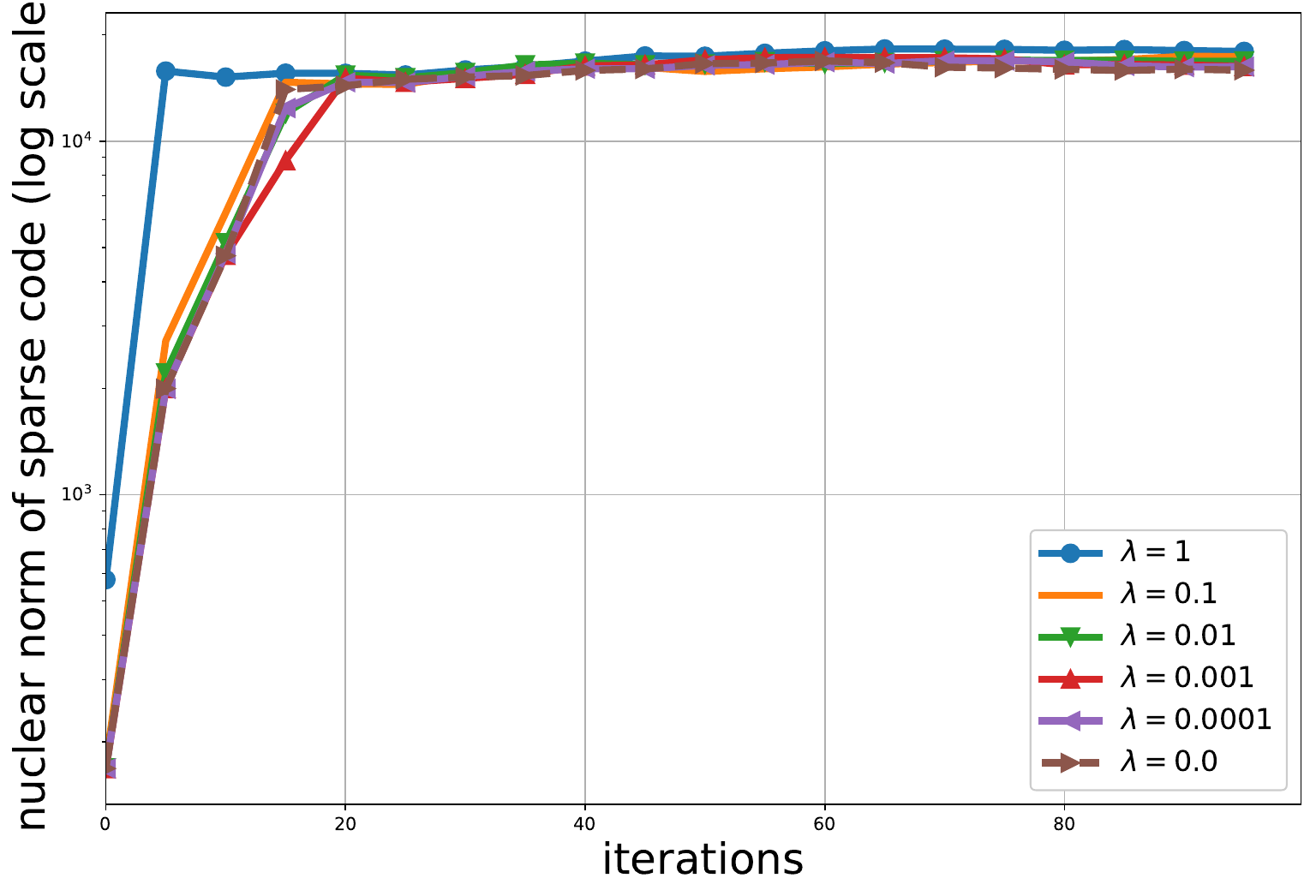}
    \caption{nuclear norm of sparse code $w$}
    \label{fig:subfig1 power}
  \end{subfigure}%
  \begin{subfigure}{.5\textwidth}
    \centering
    \includegraphics[width=1.0\linewidth]{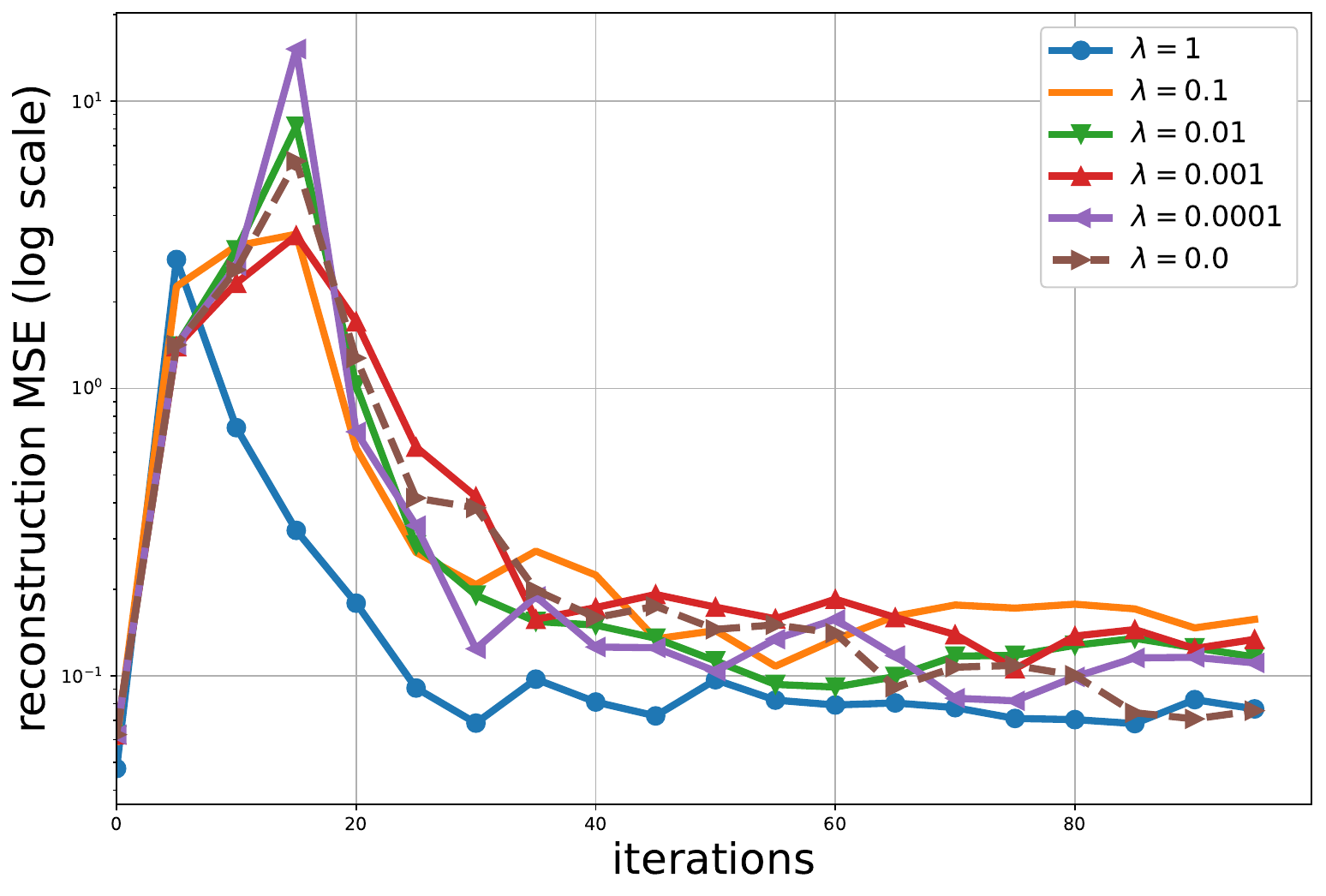}
    \caption{Reconstruction MSE of $x$}
    \label{fig:subfig2 log}
  \end{subfigure}
  \caption{Results for sparse coding reparameterisation $g(w)=\log  u - \log  v$}
  \label{fig:sparse code log}
\end{figure}

In this context, we reparameterize the sparse code as $g(w)=\log  (u) - \log  (v)\in\mathbb{R}^{n}$ and replace the regularization as discussed.
We initialize the parameters as $u_0 = {1}/{(\beta(1+e^{-x}))}$ and $v_0 ={1}/{(\beta(1+e^{x}))}$, where $\beta=1$ and $x=0.1$. 
Note that the initialization is different for stability reasons.
We explore various values for $\alpha \in \{0.0001, 0.001, 0.01, 0.1, 0.0, 1.0\}$. 
During the training process, we track two key metrics: the reconstruction Mean Squared Error (MSE) and the nuclear norm of the sparse code, defined as $g(w) = \log{u} - \log{v}$.
The results are illustrated in Figure.\ref{fig:sparse code log}. We observe that higher regularization leads to a faster increase in the nuclear norm, which confirms the movement to $L_2$ regularization. 
This leads to a construction error.

\newpage

\section{Hyperparameters and additional figures on quadratic reparameterizations}\label{appendix acc}
We present the experimental details and additional figures such as validation error and accuracy. Moreover, we present the evolution of the time-dependent Legendre functions corresponding to $m \odot w$ in Figures.\ref{fig: Breg evol mw} and \ref{fig: Breg evol ent} .
In the case of attention, we used the optimizer AdamW with a learning rate $1e-4$ and a constant learning rate.
We start training from a pretrained tiny-ViT on ImageNet.
Moreover, the CIFAR10 experiment is run over $3$ seeds.
Finally, for LoRA we use SGD with momentum ($0.9$), constant learning rate $2e-4$, LoRA rank $8$, alpha $8$ and no drop-out.
The GPT-2 experiment is run over $2$ seeds.

The validation accuracy is given in Table \ref{tab:appendix acc att} for the attention experiment.
The validation loss is given in Table \ref{tab:appendix acc} for the LoRA experiment.
In all cases turning off the weight decay leads to an improved validation score (accuracy or error). In Table \ref{tab: ablation vit imagenet} we provide an ablation for various turn off points of the weight decay for strength $0.2$. Turning off leads to better generalization at the intersection point or even before intersection.

\begin{table}[h!]
\caption{Validation accuracy and ratio at different weight decay turn-off points.}\label{tab: ablation vit imagenet}
\centering
\begin{tabular}{c|c|c|c|c}
WD Off & Intersect & Val Acc (WD) & Val Acc (Off) & Norm Ratio \\
\hline
50  & 104  & 70.7           & 72.4           & 6.9 \\
100 & 195  & 70.2           & 72.6           & 6.5 \\
150 & 270  & 71.1           & 73.4           & 6.4 \\
200 & None & 70.3 (6.3)     & 72.5 (6.1)     & -   \\
\end{tabular}
\end{table}

\begin{figure}[!htb]
  \centering
    \begin{subfigure}[b]{0.45\textwidth}
    \centering
    \includegraphics[width=0.8\textwidth]{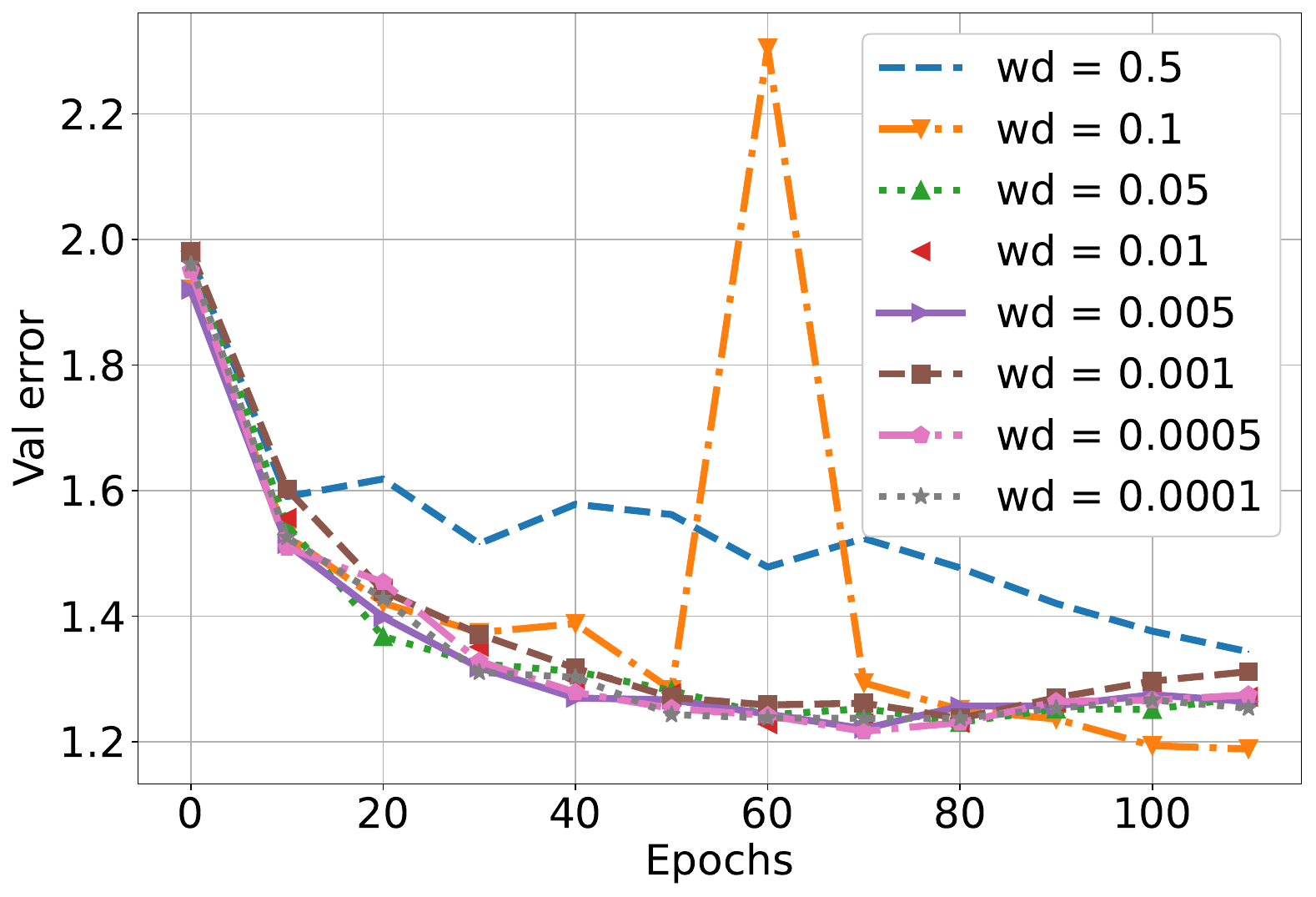}
    \caption{Validation error.}
     \label{fig:transformer_error}
    \end{subfigure}
    \hfill
    \begin{subfigure}[b]{0.45\textwidth}
    \centering
    \includegraphics[width = 0.8 \textwidth]{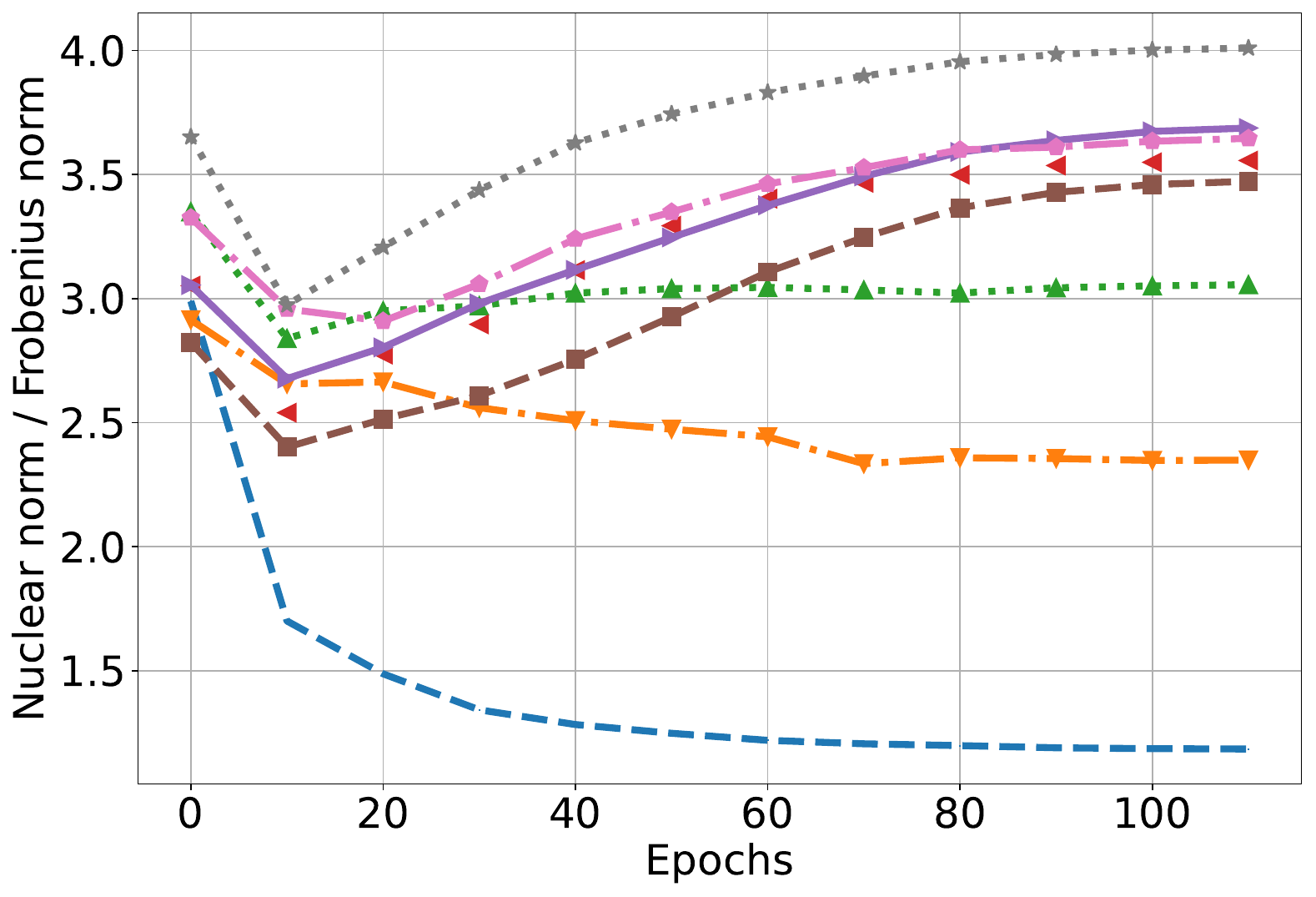}
    \caption{Ratio of the nuclear norm and the Frobenius norm.}
    \label{fig:ratio}
    \end{subfigure}
    \caption{Varying the weight decay parameter for a ViT on CIFAR10. Higher weight decay leads to a lower ratio and can also lead to lower validation errors.}
    \label{fig: attetion}
\end{figure}

Furthermore, we provide training of a tiny-ViT from scratch on CIFAR10 with varying weight decay in Figure. \ref{fig: attetion}. The learning rate is $1e-3$ and we use cosine warmup. 
Observe that higher weight decay is necessary to keep the ratio down at the end of the training.
This can improve the validation error significantly as observed for $wd = 0.1$.

\begin{table}
\caption{Validation accuracy for tiny-ViT experiment (attention).}
    \centering
    \begin{tabular}{l |ccc|ccc}
    Dataset& $wd = 0.2$ & $wd = 0.2$, turn-off & $wd = 0.1$ & $wd = 0.02$ & $wd = 0.02$, turn-off & $wd = 0.01$ \\ \hline
       ImageNet & $71.08$ & $\mathbf{73.66}$ & $71.76$ & $74.576$ & $\mathbf{75.24}$ & $75.05$ \\ 
       CIFAR10  & $52.38(\pm 0.16)$ &	$\mathbf{56.7(\pm(0.39)}$ & $54.95(\pm0.36)$ & $57.05(\pm0.27)$ & $\mathbf{57.78(\pm 0.32)}$ & $57.33(\pm0.44)$ \\ 
    \end{tabular}
    \label{tab:appendix acc att}
\end{table}

\begin{table}
\caption{Validation loss for LoRA experiment on Shakespeare dataset.}
    \centering
    \begin{tabular}{l|ccc|ccc}
    Architecture & $wd = 1.0$ & $wd = 1.0$, turn-off & $wd = 0.5$ & $wd = 0.2$ & $wd = 0.2$, turn-off & $wd = 0.1$ \\ \hline
       GPT2-xl  & $2.99$ & $\mathbf{2.96}$ & $2.97$ & $2.96$ & $\mathbf{2.95}$ & $2.96$ \\ 
       GPT2  & $3.44 (\pm 0.00)$ & $\mathbf{3.42(\pm 0.00)}$ & $3.43(\pm 0.00)$ & $3.43 (\pm 0.01)$ & $\mathbf{3.41 (\pm 0.00)}$ & $3.42(\pm 0.01)$ \\ 
    \end{tabular}
    \label{tab:appendix acc}
\end{table}

\begin{table}
\caption{Quadratic reparameterization.}
    \centering
    \begin{tabular}{c|c|c}
    Matrix sensing & $UU^T$ & $U \in \mathbb{R}^{n \times n}$ \\\hline
       Attention  &  $\textit{Softmax}( QK^T) V$ & $Q,K,V \in \mathbb{R}^{n_q \times d}$\\ \hline
       LoRA  &   $W_0 + A B$ & $A \in \mathbb{R}^{n \times r}, B \in \mathbb{R}^{r \times n}, W_0 \in \mathbb{R}^{n \times n}$ where $r << n$\\
    \end{tabular}
    \label{tab:appendix quad param}
\end{table}


\newpage

\section{Learning rate schedule}
We further study the effect of the learning rate scheduler. Specifically, we run pre-trained ViT-tiny on ImageNet classification fine-tuning task. We set the learning rate to $1e-4$ with AdamW optimisers. We also vary the weight decay in the range $[0.001, 0.003, 0.005, 0.007, 0.01]$. Moreover, for each of the settings, we train two comparison experiments, one without a learning rate scheduler, and one with the popular CosineAnnealingWarmRestarts.

\begin{figure}[ht]
  \centering
   \begin{subfigure}[b]{0.46\textwidth}
    \includegraphics[width=0.8\textwidth]{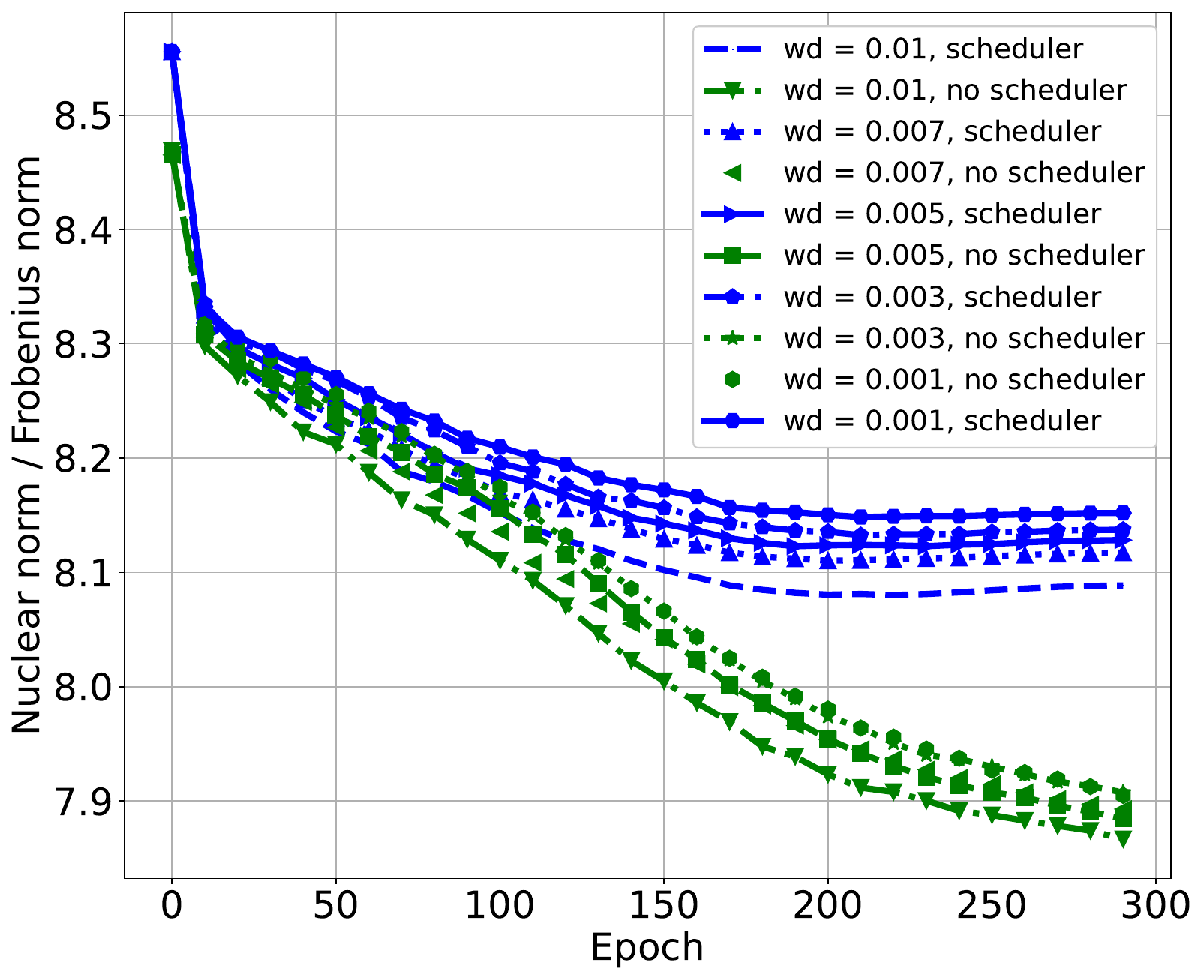}
    \caption{Average ratio ${|K^TQ|_{nuc}}/{|K^TQ|_{frob}}$}
    \label{fig:vit-no scheduler}
    \end{subfigure}
    \hfill
    \begin{subfigure}[b]{0.46\textwidth}
    \includegraphics[width =0.8\textwidth]{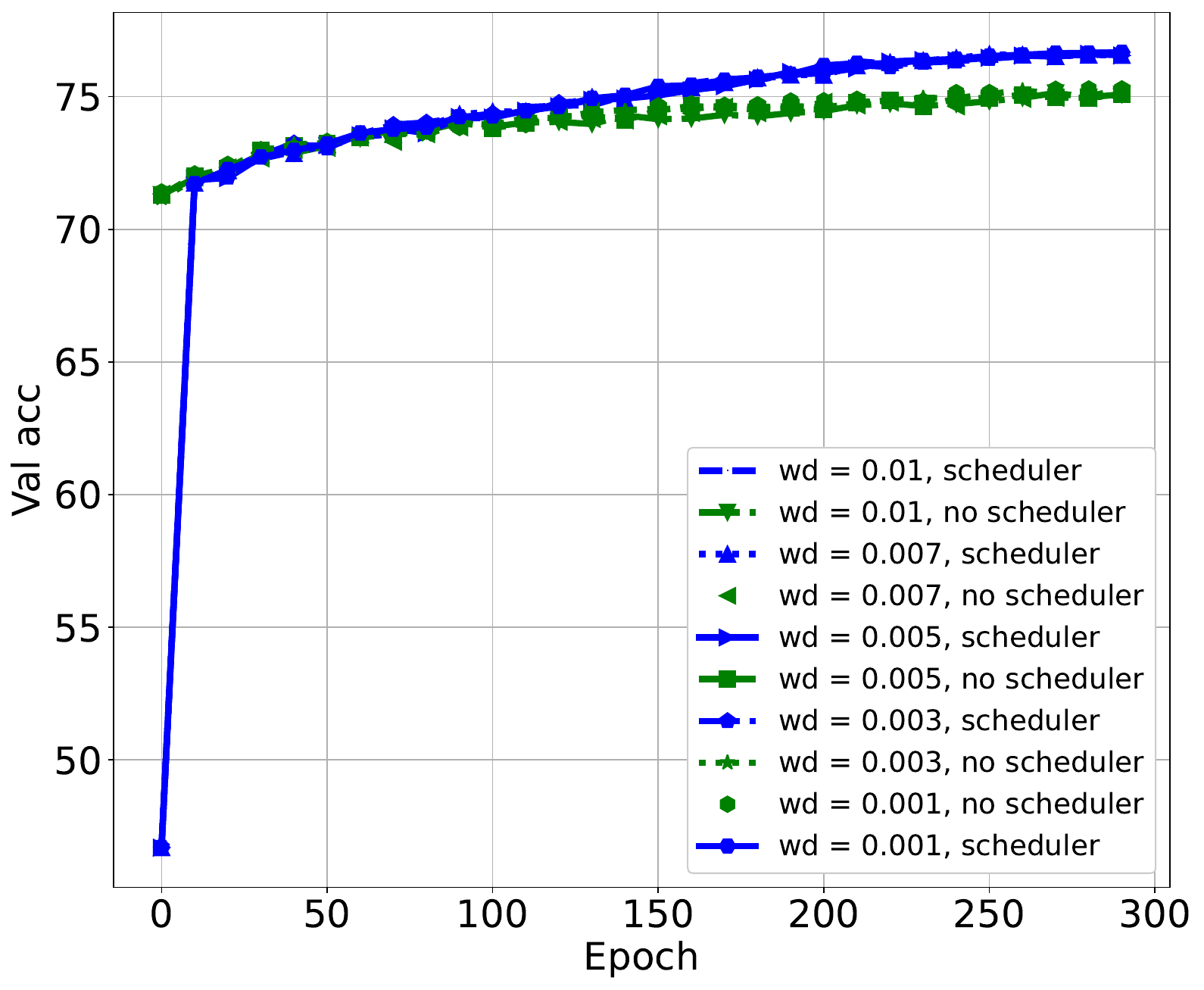}
    \caption{Validation accuracy}
    \label{fig:vit-scheduler}
    \end{subfigure}
    \caption{Results for ViT-tiny fine-tuning task with AdamW optimiser on ImageNet.}
    \label{fig: ratio comparison for vit fine-tuning}
\end{figure}

\begin{figure}[ht]
  \centering
    \begin{subfigure}[b]{0.46\textwidth}
    \includegraphics[width=\textwidth]{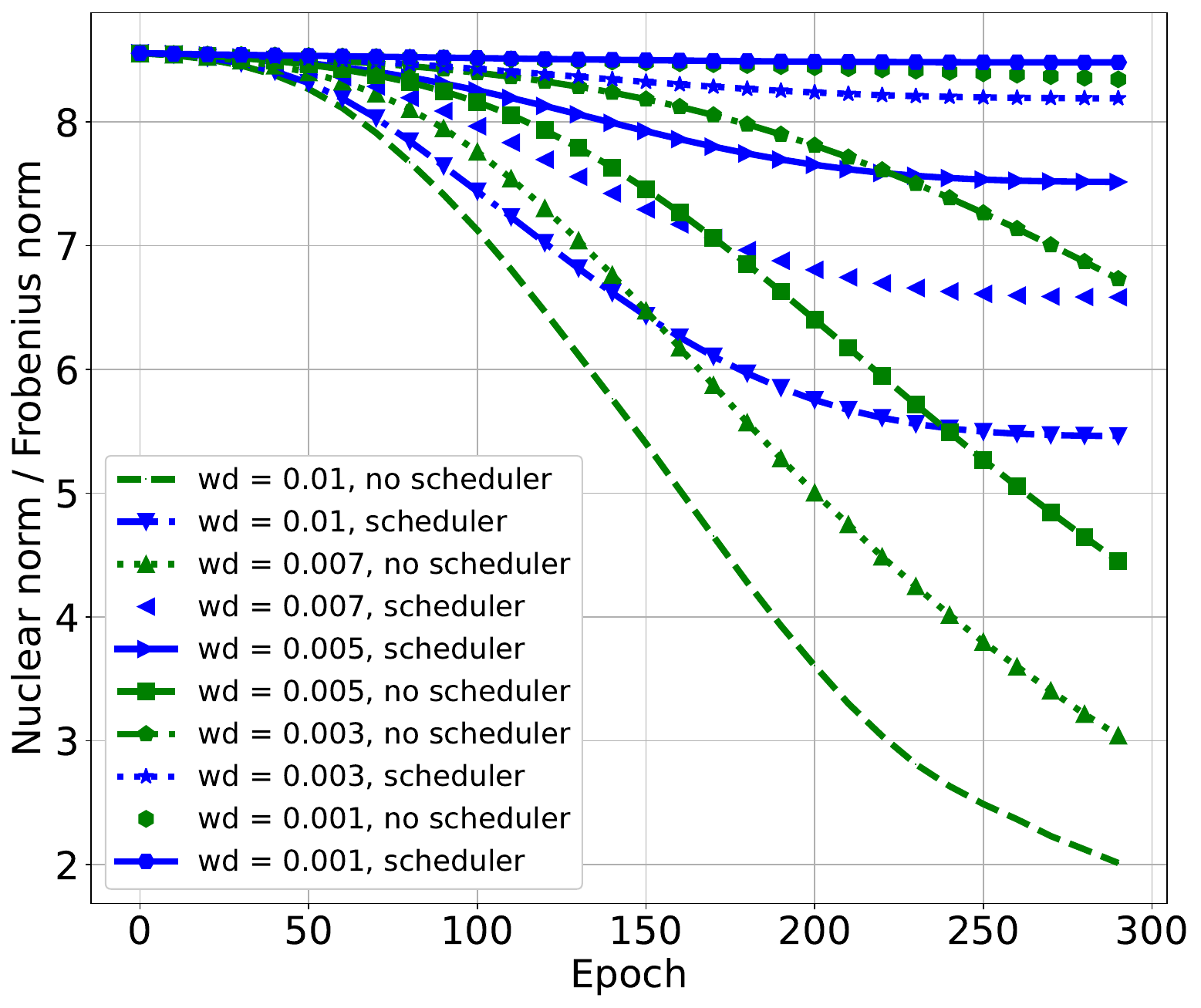}
    \caption{Average ratio ${|K^TQ|_{nuc}}/{|K^TQ|_{frob}}$}
    \label{fig:sgd vit-no scheduler}
    \end{subfigure}
    \hfill
    \begin{subfigure}[b]{0.46\textwidth}
    \includegraphics[width = \textwidth]{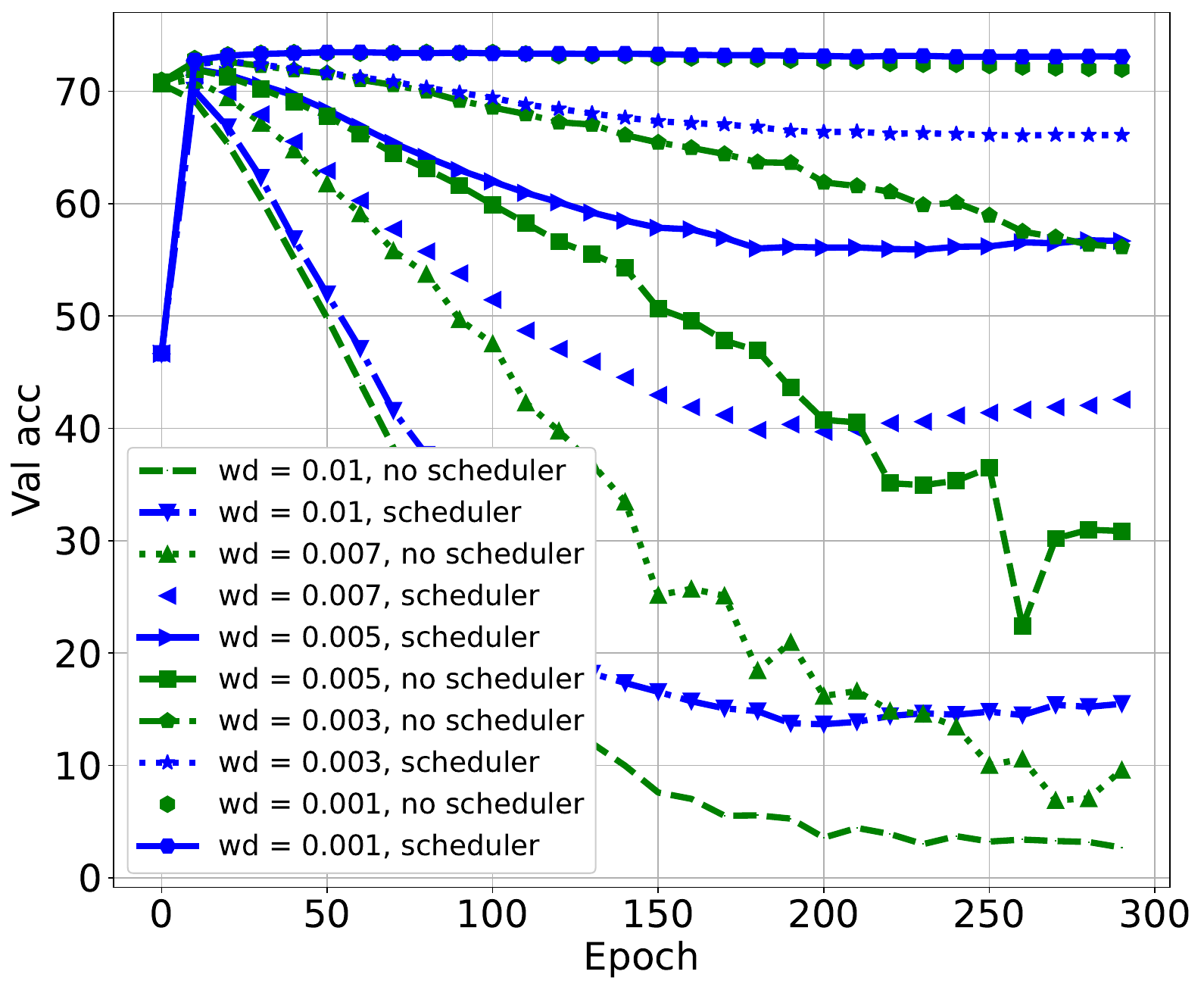}
    \caption{Validation accuracy}
   \label{fig:sgd vit-scheduler}
    \end{subfigure}
    \caption{Results for ViT-tiny fine-tuning task with SGD optimiser on ImageNet.}
    \label{fig: ratio comparison for vit SGD fine-tuning}
\end{figure}

The results are shown in Figure.\ref{fig: ratio comparison for vit fine-tuning}. Furthermore, results with SGD optimizer are included in Figure.\ref{fig: ratio comparison for vit SGD fine-tuning}.
We observe in both figures that the validation accuracy increases for the decaying schedule in comparison to the constant schedule. Moreover, we again observe a decaying ratio, for stronger weight decay the ratio decreases more. 
Moreover, decaying the learning rate has a similar effect as turning off the weight decay on the implicit and explicit regularization. Note that the AdamW optimizer can have additional effects that also contribute to changing the ratio.
Furthermore, regardless of the learning rate schedule, the ratio is decreasing indicating a modulation from Frobenius norm towards nuclear norm minimization.

\end{document}